\documentclass[twoside,11pt,preprint]{article}

\usepackage{jmlr2e}

\usepackage{microtype}
\usepackage{booktabs,bbm,amsmath,tabularx,inputenc}
\usepackage{xcolor,bm,amsmath,graphicx,comment,bigints}
\usepackage{algorithm,algorithmic,caption,subcaption,xspace}

\newcommand\cvx{$\mathcal{H}^{tr}$\xspace}
\newcommand\dtr{$\mathcal{D}^{tr}$\xspace}

\UseRawInputEncoding
\jmlrheading{1}{2000}{1-48}{4/00}{10/00}{meila00a}{Yousefzadeh}

\ShortHeadings{Deep Learning Generalization and the Convex Hull of Training Sets}{Yousefzadeh}
\firstpageno{1}

\begin{document}

\title{Deep Learning Generalization\\ and the Convex Hull of Training Sets}

\author{\name Roozbeh Yousefzadeh \email roozbeh.yousefzadeh@yale.edu \\
       \addr Yale University and VA Connecticut Healthcare System\\
       New Haven, CT 06510, USA
    %   }
%      \AND
%      \name Dianne P. O'Leary \email oleary@umd.edu \\
%      \addr Computer Science Department and Institute for Advanced Computer Studies,\\
%      University of Maryland, College Park, MD 20742, USA
      }

%\editor{Kevin Murphy and Bernhard Sch{\"o}lkopf}

\maketitle

\begin{abstract}%   <- trailing '%' for backward compatibility of .sty file
We study the generalization of deep learning models in relation to the convex hull of their training sets. A trained image classifier basically partitions its domain via decision boundaries and assigns a class to each of those partitions. The location of decision boundaries inside the convex hull of training set can be investigated in relation to the training samples. However, our analysis shows that in standard image classification datasets, all testing images are considerably outside that convex hull, in the pixel space, in the wavelet space, and in the internal representations learned by deep networks. Therefore, the performance of a trained model partially depends on how its decision boundaries are extended outside the convex hull of its training data. From this perspective which is not studied before, over-parameterization of deep learning models may be considered a necessity for shaping the extension of decision boundaries. At the same time, over-parameterization should be accompanied by a specific training regime, in order to yield a model that not only fits the training set, but also its decision boundaries extend desirably outside the convex hull. To illustrate this, we investigate the decision boundaries of a neural network, with various degrees of parameters, inside and outside the convex hull of its training set. Moreover, we use a polynomial decision boundary to study the necessity of over-parameterization and the influence of training regime in shaping its extensions outside the convex hull of training set.
\end{abstract}

\begin{keywords}
  Deep learning, Convex hulls, Generalization
\end{keywords}

\section{Introduction}

A deep learning image classifier is a mathematical function that maps images to classes, i.e., a deep learning function \citep{strang2019linear}. These models/functions have shown to be exceptionally useful in real-world applications. However, generalization of these functions is considered a mystery by deep learning researchers~\citep{arora2019implicit}. These models have orders of magnitude more parameters than their training samples \citep{belkin2019reconciling,neyshabur2019towards}, and they can achieve perfect accuracy on their training sets, even when the training images are randomly labeled, or the contents of images are replaced with random noise \citep{zhang2016understanding}. The training loss function of these models has infinite number of minimizers, where only a small subset of those minimizers generalize well \citep{neyshabur2017exploring}. If one succeeds in picking a good minimizer of training loss, the model can classify the testing images correctly, nevertheless, for any correctly classified image, there are infinite number of images that look the same, but models will classify them incorrectly (phenomenon known as adversarial vulnerability) \citep{papernot2016limitations,shafahi2018adversarial,tsipras2018robustness}. Here, we study some geometric properties of standard training and testing sets to provide new insights about what a model can learn from its training data, and how it can generalize.

Specifically, we study the convex hulls of image classification datasets (both in the pixel space and in the wavelet space), and show that all testing images are considerably outside the convex hull of training sets, with various distances from the hull. Investigating the representation of images in the internal layers of a trained CNN also reveal that testing samples are generally outside the convex hull of their training sets. We investigate the perturbations required to bring the testing images to the surface of convex hull and observe that the directions to convex hulls contain valuable information about images, corresponding to distinctive features in them. We also see that reaching the convex hull significantly affects the contents of images. Therefore, the performance of a trained model partially depends on how well it can extrapolate. We investigate this extrapolation in relation to the over-parameterization of neural networks and the influence of training regimes in shaping the extensions of decision boundaries.

% focus on the decision boundaries of the models outside the convex hulls. We show that a considerable portion of testing images are outside the convex hull of training sets. 

% , but in close proximity of convex hull boundary. 

% A small portion of testing images that are far outside the convex hull.
% when testing data is in close proximity of the convex hull of training set, the decision boundaries of the deep learning function can be a reliable extension of what the model has learned from the data. However, in certain neighborhoods outside the convex hulls, the decision boundaries can be completely random. We relate our observations to the tangible case of polynomial fitting and explain that part of the deep learning generalization can be explained by the training methods designed based on the knowledge of testing data.

\subsection{Related work about geometry of data and deep learning}

Convex hull of training sets are not commonly considered in deep learning studies, especially the ones focused on their generalization. Recently, \citep{yousefzadeh2020wspectral} reported that in a reduced (100 dimensional) wavelet space, distance of testing images to the convex hull for each training class can predict the label for more than 98.5\% of MNIST testing data. Previously, \citep{haffner2002escaping} considered the convex hull of MNIST data for Support Vector Machines. Similarly, \citet{vincent2002k} considered the convex hulls for K-Nearest Neighbor (KNN) algorithms. However, those methods do not generalize to deep learning functions.

Some researchers have studied other geometrical aspects of deep learning models, e.g., \citep{cohen2020separability,fawzi2018empirical,cooper2018loss,kanbak2018geometric,neyshabur2017geometry}. However, those studies do not investigate the generalization of deep neural networks in relation to the convex hull of training sets. Most recently, \citet{xu2020neural} studied the extrapolation behavior of ReLU perceptrons and concluded that such models cannot extrapolate most non-linear tasks. However, they do not connect their analysis to the fact that a considerable portion of testing samples of standard image datasets fall outside the convex hull of their training sets.

There are less recent studies on neural networks that consider their extrapolation capability. For example, \citet{barnard1992extrapolation} explains poor performance of neural networks on some tasks by considering them as extrapolation outside what the model has learned. Also, \citet{psichogios1992hybrid} reports that neural networks do not perform well when the data is not within their training data range. Later, \citet{kosanovich1996improving} tried to improve the extrapolation capability of neural networks for model-based control in chemical processes.

\subsection{Our plan}

In Section~\ref{sec:geometry}, we study geometric properties of standard image-classification datasets and how testing samples relate to the convex hull of training sets. In Section~\ref{sec:polynomial}, we consider a classification task in 2-dimensional space using a polynomial decision boundary. We study how the extensions of the polynomial decision boundary can be reshaped for extrapolation, and how that relates to over-parameterize and training regime. Section~\ref{sec:deep_output} investigates how the output of trained networks can vary in the under-parameterized and over-parameterized regimes. Finally, Section~\ref{sec:conclusion} presents our conclusions and describes the avenues for future research.

\section{Geometry of testing data w.r.t the convex hull of training sets} \label{sec:geometry}

First, we show that for standard datasets: MNIST \citep{lecun1998gradient} and CIFAR-10 \citep{krizhevsky2009learning}, most of their testing data are considerably outside the convex hull of their training sets. We denote the convex hull of a training set by $\mathcal{H}^{tr}$.

To verify whether an image/data point is inside its corresponding $\mathcal{H}^{tr}$ or not, we can simply try to fit a hyper-plane separating the point and the training set. If we find such hyper-plane, the point is outside the convex hull, and vice versa. This is basically a linear regression problem and there are many efficient and fast methods to perform it, e.g., \citep{goldstein2015adaptive}. For both MNIST and CIFAR-10 datasets, all testing images are outside the $\mathcal{H}^{tr}$, in the pixel space and in the wavelet space.% For CIFAR-10, that percentage is more than 99.9\%.% When we transform the images with wavelets (an operation analogous to convolutional neural networks), these percentages almost remain the same.

% Figure~\ref{fig:conv_perc} shows the percentage of testing data inside and outside the $\mathcal{H}^{tr}$ for MNIST and CIFAR-10 datasets.

% \begin{figure}[h]
%   \centering
%   \includegraphics[width=0.2\linewidth]{figures/}
%   \caption{Percentage of testing images inside and outside the convex hull of training set, $\mathcal{H}^{tr}$, for (a) MNIST and (b) CIFAR-10 datasets.}
%   \label{fig:conv_perc}
% \end{figure}

We then investigate the testing data outside the $\mathcal{H}^{tr}$, to see how far they are located from it. For every testing image, $x^{te}$, that is outside the \cvx, we would like to find the closest point to it on the \cvx, and we denote that point by $x^{\mathcal{H}}$. We are interested to know the direction of the shortest vector that can bring the testing image to \cvx. We are also interested in how far the testing images are from the \cvx.

\subsection{Computing the point on \cvx closest to a query point outside \cvx}

The point $x^{\mathcal{H}}$ on the \cvx, closest to a query point $x^{te}$ outside \cvx, is the solution to a well defined optimization problem. Let's consider a dataset, \dtr, formed as a matrix, with $n$ rows corresponding to the samples, and $d$ columns corresponding to the features, i.e., dimensions.% \cvx is the convex hull of all the samples in \dtr. Our query point sits outside the \cvx, and we seek to find $x$, the point in \cvx that is closest to $q$.

To ensure that $x^{\mathcal{H}}$ belongs to \cvx, we can define
\begin{equation} \label{eq:alpha}
    x^{\mathcal{H}} = \alpha \mathcal{D}^{tr},
\end{equation}

where $\alpha$ is a row vector of size $n$. If all elements of $\alpha$ are bounded between 0 and 1, and their summation also equals 1, then by definition, $x$ belongs to \cvx. Given equation~\eqref{eq:alpha}, we can change our optimization variable to $\alpha$.

Our objective function is

\begin{equation} \label{eq:obj}
    \min_{\alpha} \| x^{te} - \alpha \mathcal{D}^{tr}\|_2^2,
\end{equation}

while our constraints ensure that $x^{te}$ belongs to \cvx.

\begin{equation} \label{eq:const1}
    \alpha \mathbbm{1}_{n,1} = 1,
\end{equation}
\begin{equation} \label{eq:const2}
    0 \leq \alpha \leq 1.
\end{equation}

This is a constrained least squares problem which can be solved using algorithms in numerical optimization literature \citep{nocedal2006numerical,oleary2013variable}. For any given query point, $x^{te}$, we first compute the optimal $\alpha$ using equations \eqref{eq:obj}-\eqref{eq:const2}. We then compute the corresponding $x^{\mathcal{H}}$ using the optimal $\alpha$ and equation~\eqref{eq:alpha}.

We solve the optimization problem, numerically, using the gradient projection algorithm described by \citet[Chapter 16]{nocedal2006numerical}. Note that size of $\mathcal{D}^{tr}$ can be quite large (e.g., for CIFAR-10, it is 50,000x3,072) and solving the above optimization problem may be time consuming. To make the optimization faster, we used a sketching scheme in which the $\mathcal{D}^{tr}$ is divided into a few pieces. We first solve the problem using the first piece of $\mathcal{D}^{tr}$, then we add the second piece, and solve the problem again using the solution from previous step as the starting point. This process continues until all pieces of $\mathcal{D}^{tr}$ are included. The final solution is obtained by solving the problem for the entire $\mathcal{D}^{tr}$. The problem is convex and we verify the optimality of our solutions by satisfying the Karush Kuhn Tucker (KKT) conditions \citep{nocedal2006numerical}.\footnote{Since our objective function is a linear least squares problem and our constraints are all linear, we only need to satisfy the first order optimality conditions.} Using the above sketching scheme only makes the process faster and does not affect the optimal solutions.

%as described in Appendix~\ref{appx:alg}, and ensured the optimality of the 

We note that there are approximation algorithms that estimate the solution to this problem. In our experiments, approximation algorithm by \cite{blum2019sparse} was faster than numerical optimization and its approximate solutions were generally a good estimate of the distance to convex hull, but the resulting images from approximation algorithm were not clear enough visually. At the end, we opted for more accuracy and used numerical optimization to solve the problem.

% The objective function seeks to minimize the distance between $x^{te}_i$ and $x^{\mathcal{H}}_i$, while the linear constraints ensure that $x^{\mathcal{H}}_i \in \mathcal{H}^{tr}$. 

\subsection{How far are testing images from the \cvx?}

\begin{figure}[H]
    \centering
     \begin{subfigure}[b]{0.46\textwidth}
         \centering
         \includegraphics[width=1\linewidth]{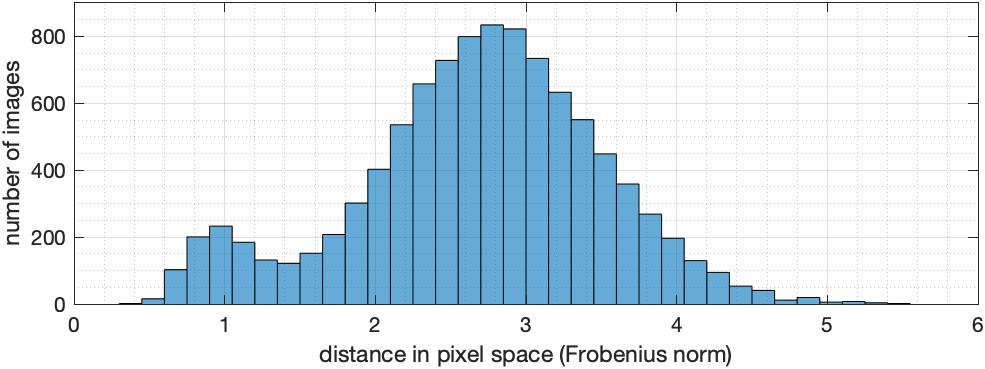}
         \caption{MNIST (pixel space)}
         \label{fig:data}
     \end{subfigure}
     \quad
     \begin{subfigure}[b]{0.46\textwidth}
         \centering
         \includegraphics[width=1\linewidth]{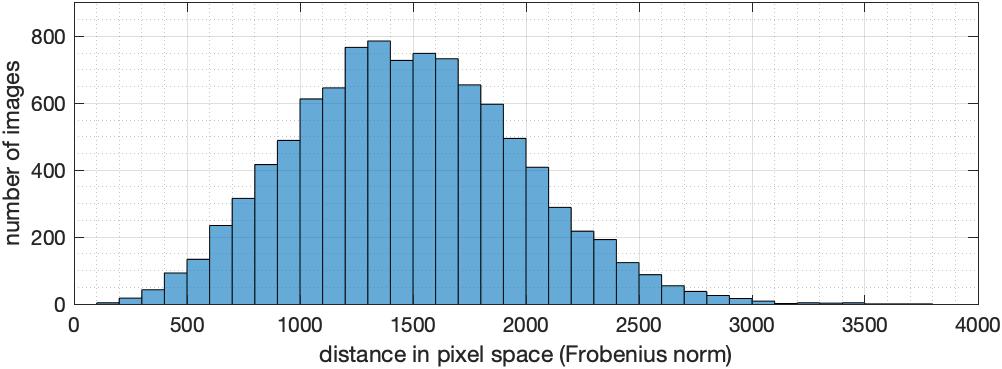}
         \caption{CIFAR-10 (pixel space)}
         \label{fig:polynomial}
     \end{subfigure}
    %  \begin{subfigure}[b]{0.4\textwidth}
    %      \centering
    %      \includegraphics[width=1\linewidth]{figures/dist_w_mnist.jpg}
    %      \caption{MNIST (wavelet space)}
    %      \label{fig:data}
    %  \end{subfigure}
    %  \quad
    %  \begin{subfigure}[b]{0.4\textwidth}
    %      \centering
    %      \includegraphics[width=1\linewidth]{figures/dist_w_cifar10.jpg}
    %      \caption{CIFAR-10 (wavelet space)}
    %      \label{fig:polynomial}
    %  \end{subfigure}   
    \caption{Variations of distance to $\mathcal{H}^{tr}$ for all testing images.}
  \label{fig:conv_dist}
\end{figure}

\begin{figure}[H]
    \centering
     \begin{subfigure}[b]{0.48\textwidth}
         \centering
         \includegraphics[width=.18\linewidth]{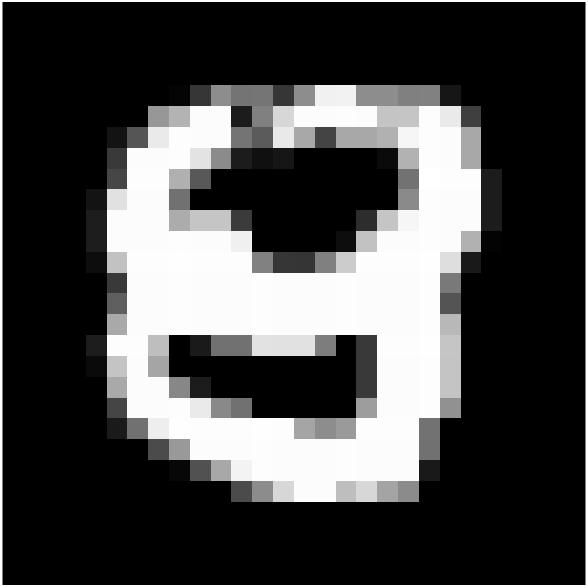}
         \includegraphics[width=.18\linewidth]{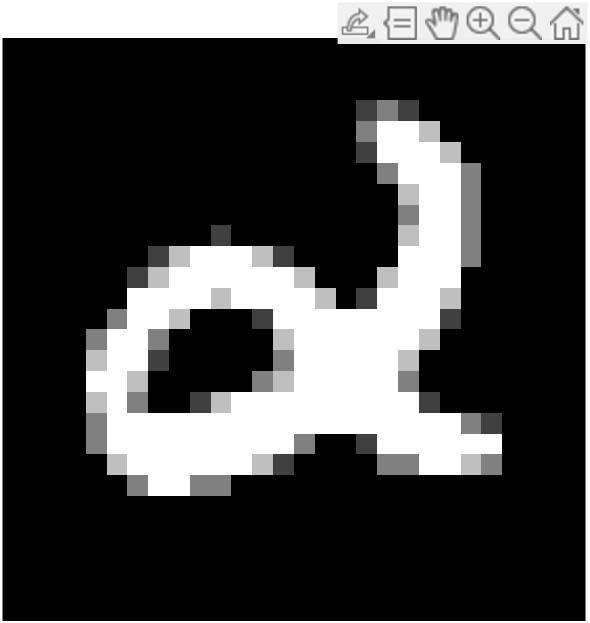}
         \includegraphics[width=.18\linewidth]{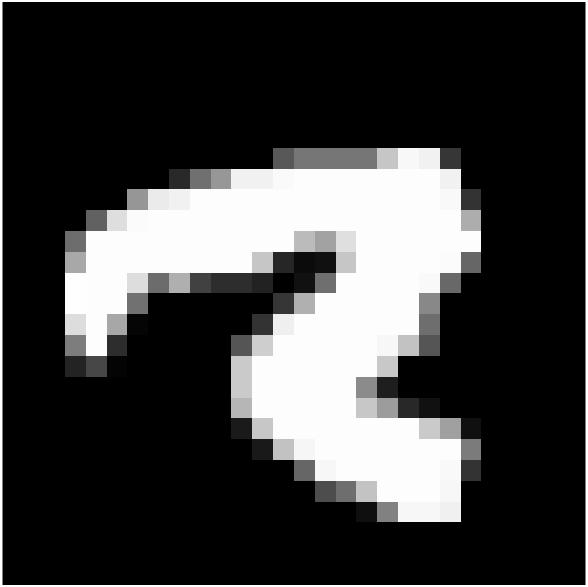}
         \includegraphics[width=.18\linewidth]{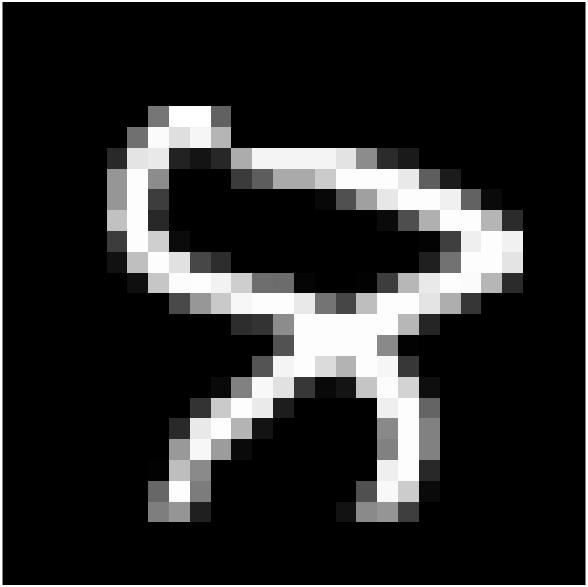}
         \includegraphics[width=.18\linewidth]{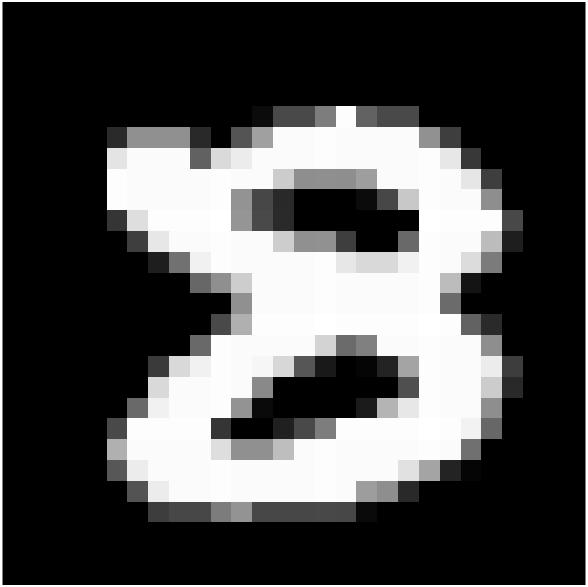}
         \caption{MNIST}
         \label{fig:far_mnist}
     \end{subfigure}
     \quad
     \begin{subfigure}[b]{0.48\textwidth}
         \centering
         \includegraphics[width=.18\linewidth]{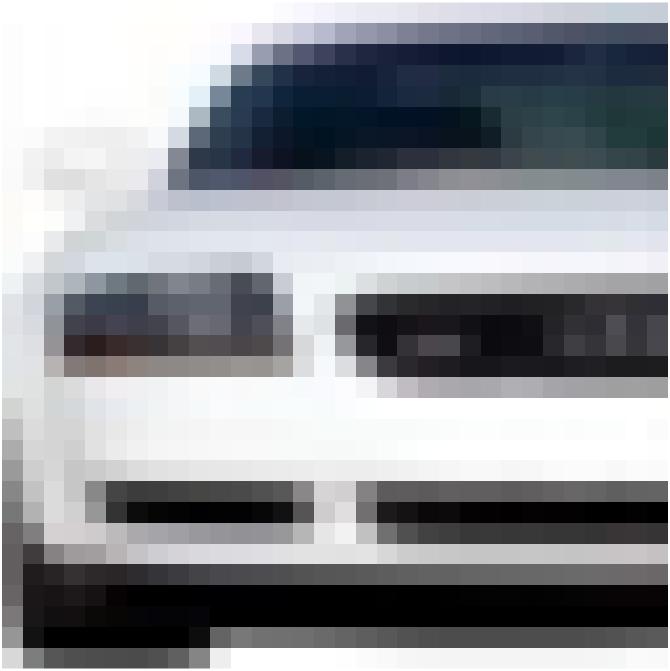}
         \includegraphics[width=.18\linewidth]{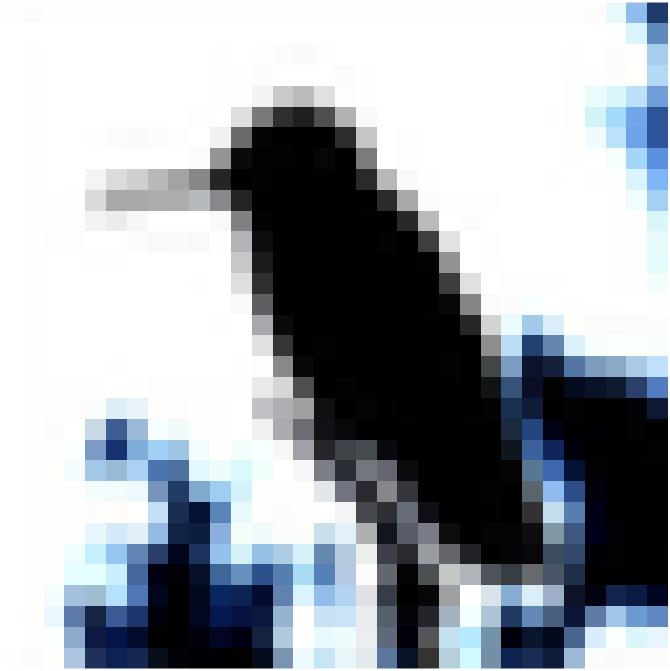}
         \includegraphics[width=.18\linewidth]{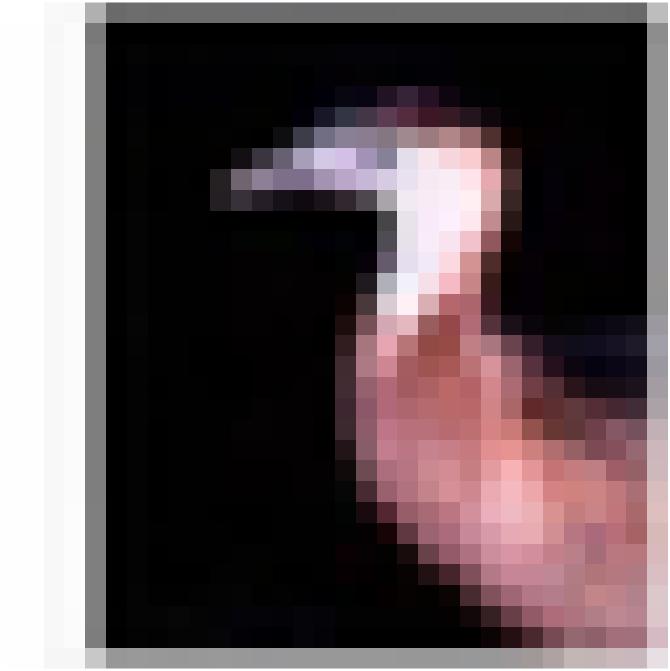}
         \includegraphics[width=.18\linewidth]{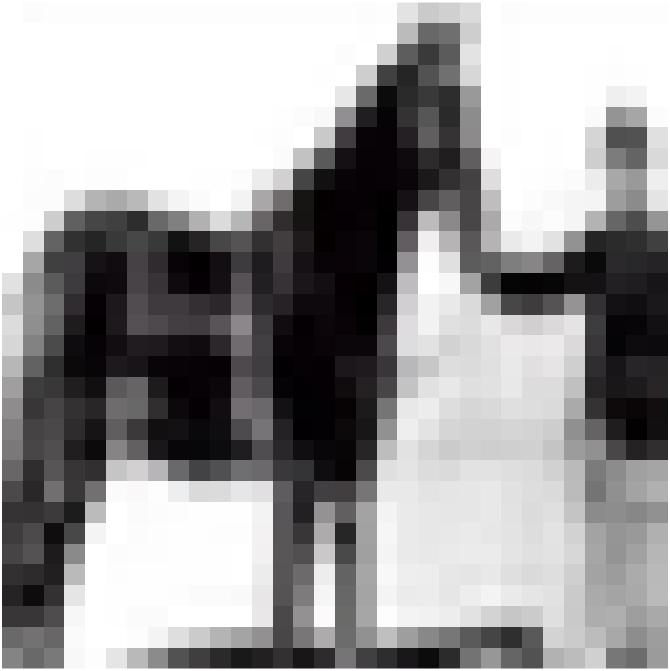}
         \includegraphics[width=.18\linewidth]{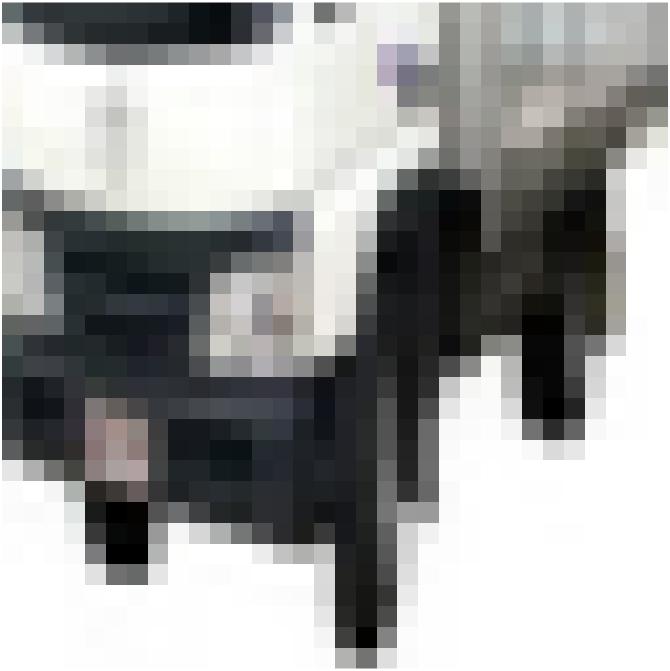}
         \caption{CIFAR-10}
         \label{fig:far_cif10}
     \end{subfigure}
    \caption{Samples of testing images that are farthest from their \cvx.}
  \label{fig:farthest}
\end{figure}

Using the optimization problem described in previous section, we compute the distance of testing samples from the \cvx, for the MNIST and CIFAR-10 datasets. Figure~\ref{fig:conv_dist} shows the histogram of distances and Figure~\ref{fig:farthest} shows samples of testing images that are farthest from their \cvx.

% Here, we approximate the distance to convex hull by first fitting a linear Support Vector Machine (SVM) between the testing point and the $\mathcal{H}^{tr}$. Since an SVM maximizes its margin from its supports, the total margin of the resulting SVM will closely approximate the distance between the point and $\mathcal{H}^{tr}$\footnote{In our experiments, we observe that in most cases, our computed SVMs are equidistant (or almost equidistant) from the testing points and the closest point of $\mathcal{H}^{tr}$. We note that this approximation of distance (i.e., using a linear SVM) does not overestimate the distance to \cvx. In fact, the actual distances can be larger than the ones we report.}. 

To get a better sense of how far these distances are, consider the \cvx of CIFAR-10 dataset. Its diameter, the largest distance between any pair of vertices in \cvx, is 13,621 (measured by Frobenius norm in pixel space). On the other hand, the distance of farthest testing image from the \cvx is about 3,500 (about 27\% of the diameter of \cvx). Moreover, the average distance between pairs of images in the training set of CIFAR-10 is 4,838, while the closest pair of images are only 701 apart.

%In other words, the distance of testing data to \cvx can be more than 1/10th of the diameter of training set. 
Hence, the distance of testing data to \cvx is not negligible and we cannot dismiss it as a small noise. However, it is not very large either. When we convolve the images with Haar or Daubechies \citep{daubechies1992ten} wavelets (an operation analogous to convolutions performed by CNNs), the testing samples still remain outside the \cvx in the wavelet space. When we choose a small subset of wavelet coefficients for all images, still, the testing samples remain outside the \cvx in that lower dimensional space, and the distances still resemble a normal distribution.

Overall, we can say that in order to classify most of the testing images in the above datasets, a model has to extrapolate, to some moderate degree, outside its \cvx. In Appendix~\ref{appx:random_points}, we discuss the convex hull of random points in high-dimensional space and explain that the distance to convex hulls will be much larger if we deal with random points instead of these images. This signifies that training and testing sets of these datasets are related to each other, and the extent of extrapolation task is limited.

\subsection{Directions to \cvx and the information they contain}

\begin{figure}[H]
\begin{center}
    % \begin{minipage}[b]{1\columnwidth}
    % \centering
    % (original) \raisebox{-.5\height}{\includegraphics[width=.15\columnwidth]{figures/mnist_te2119orig.jpg}} {\vspace{.2cm} -}
    % \raisebox{-.5\height}	{\includegraphics[width=.15\columnwidth]{figures/mnist_te2119diff_all.jpg}} =
    % \raisebox{-.5\height}	{\includegraphics[width=.15\columnwidth]{figures/mnist_te2119oncvx_all.jpg}} (on \cvx)
    % \end{minipage}
    \begin{minipage}[b]{.85\columnwidth}
    \centering
    (original) \raisebox{-.5\height}{\includegraphics[width=.12\columnwidth]{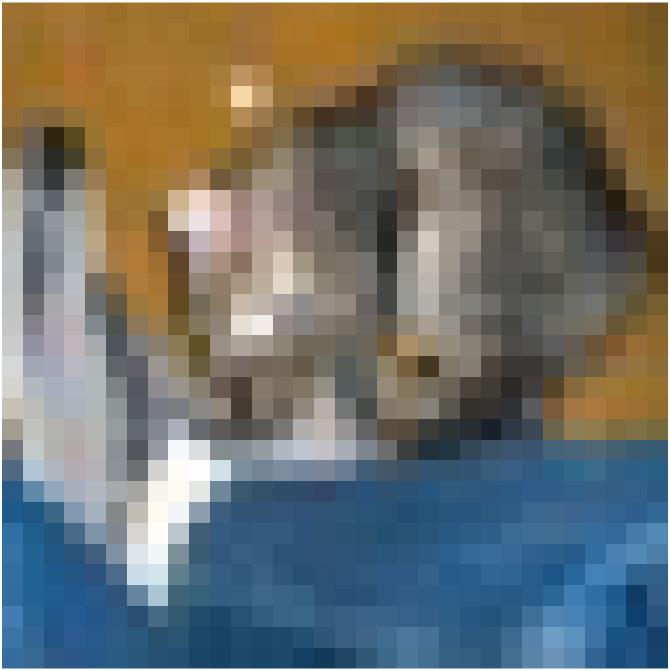}} {\vspace{.2cm} -}
    \raisebox{-.5\height}	{\includegraphics[width=.12\columnwidth]{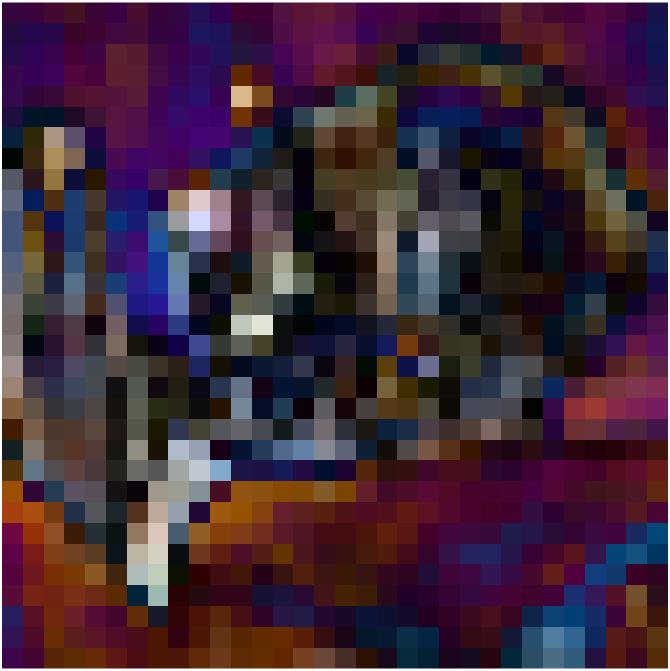}} =
    \raisebox{-.5\height}	{\includegraphics[width=.12\columnwidth]{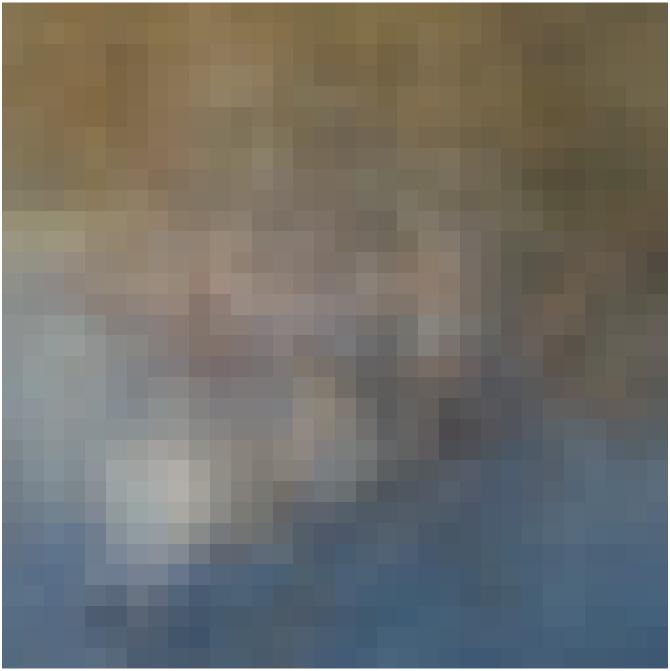}} (on \cvx)
    \end{minipage}
    \begin{minipage}[b]{.85\columnwidth}
    \centering
    (original) \raisebox{-.5\height}{\includegraphics[width=.12\columnwidth]{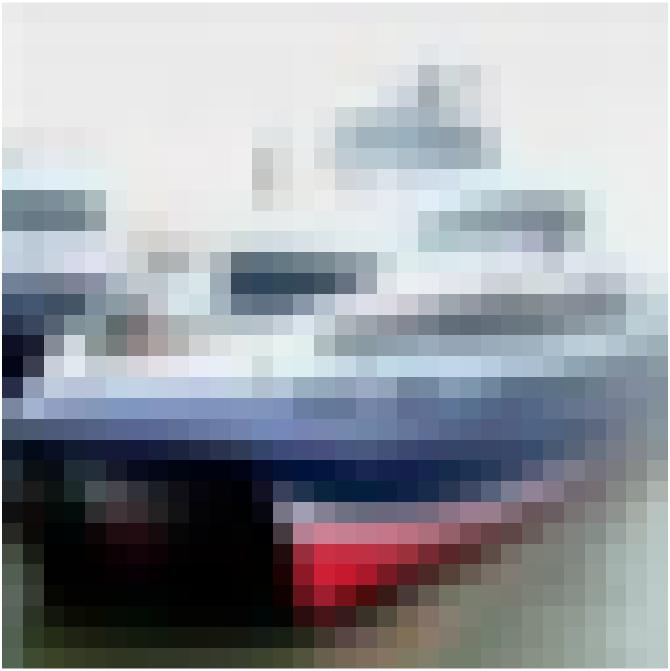}} {\vspace{.2cm} -}
    \raisebox{-.5\height}	{\includegraphics[width=.12\columnwidth]{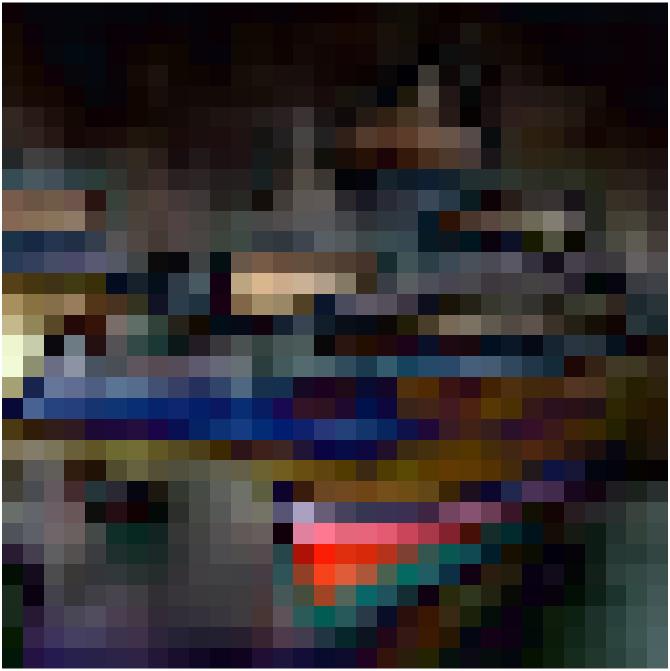}} =
    \raisebox{-.5\height}	{\includegraphics[width=.12\columnwidth]{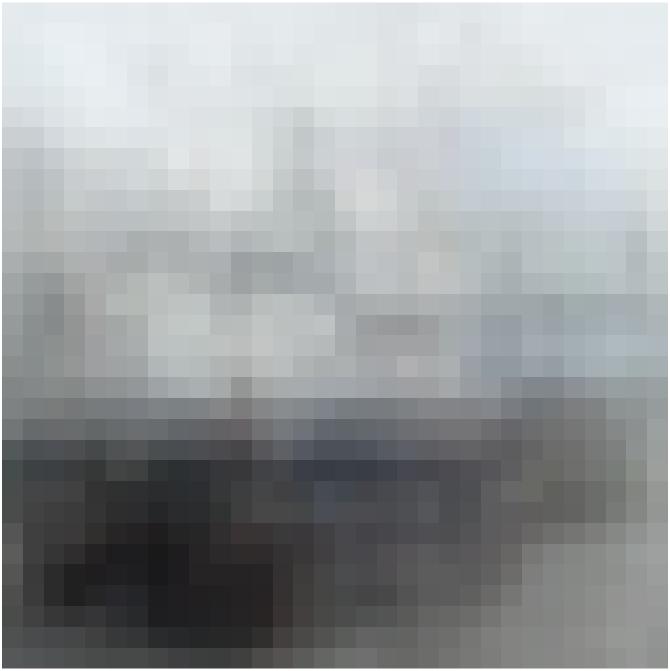}} (on \cvx)
    \end{minipage}
    \begin{minipage}[b]{.85\columnwidth}
    \centering
    (original) \raisebox{-.5\height}{\includegraphics[width=.12\columnwidth]{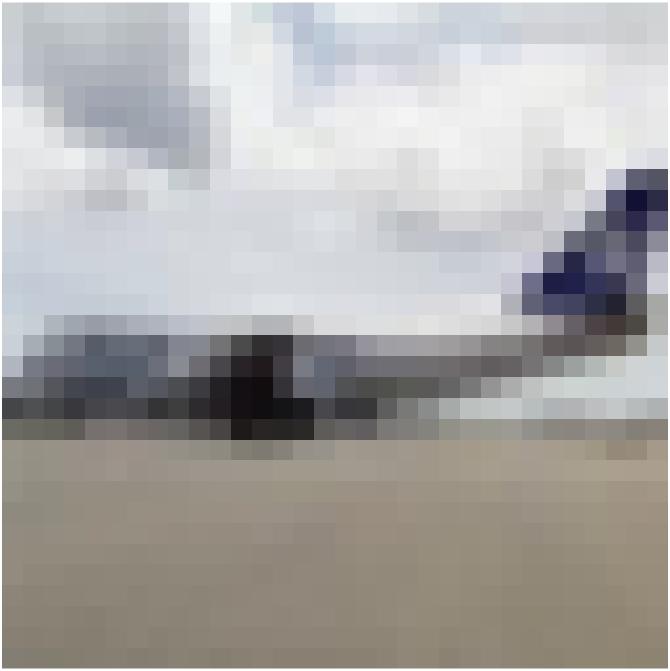}} {\vspace{.2cm} -}
    \raisebox{-.5\height}	{\includegraphics[width=.12\columnwidth]{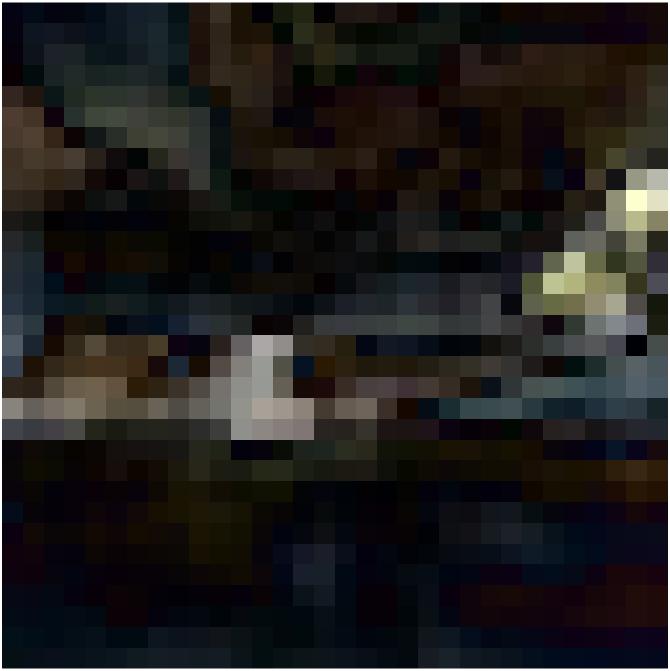}} =
    \raisebox{-.5\height}	{\includegraphics[width=.12\columnwidth]{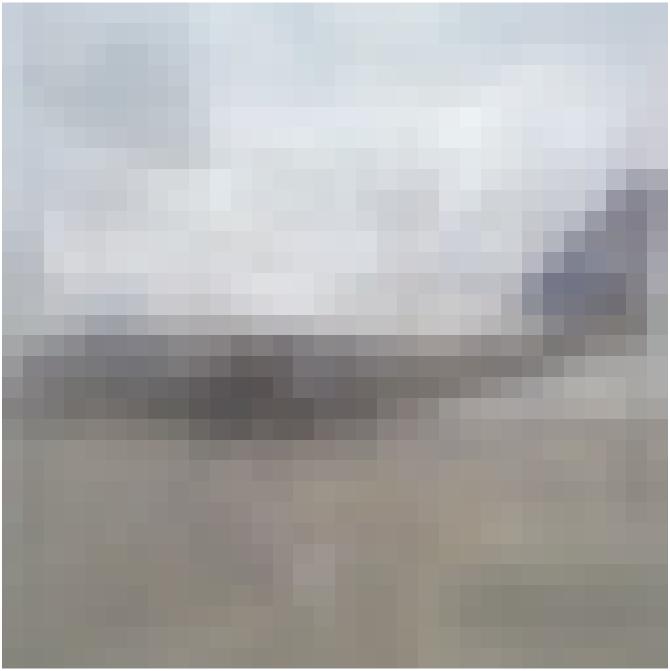}} (on \cvx)
    \end{minipage}
\end{center}
    \caption{Minimum perturbation that can bring a testing image to \cvx of all classes. (left image) original testing image from CIFAR-10, (middle image) what should be removed from the original image, (right image) the resulting image on the \cvx. These directions specifically relate to the objects of interest depicted in images.}
  \label{fig:conv_perturb_cif10}
\end{figure}

\begin{figure}[h]
\begin{center}
    \begin{minipage}[b]{.64\columnwidth}
    \centering
    \raisebox{-.4\height}{\includegraphics[width=.15\columnwidth]{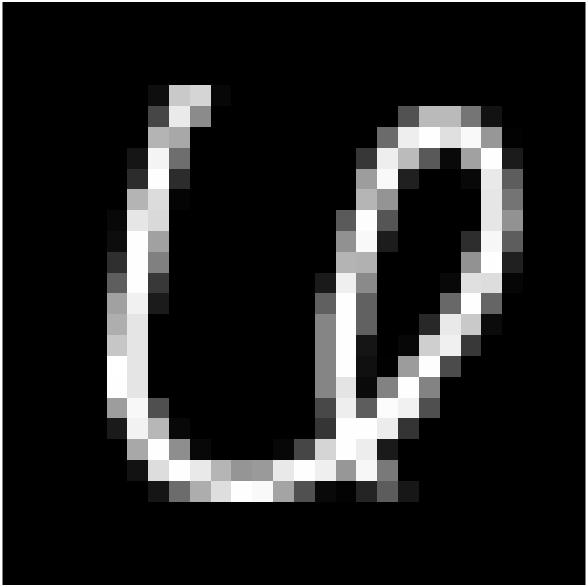}} {\vspace{.2cm} -}
    \raisebox{-.4\height}	{\includegraphics[width=.15\columnwidth]{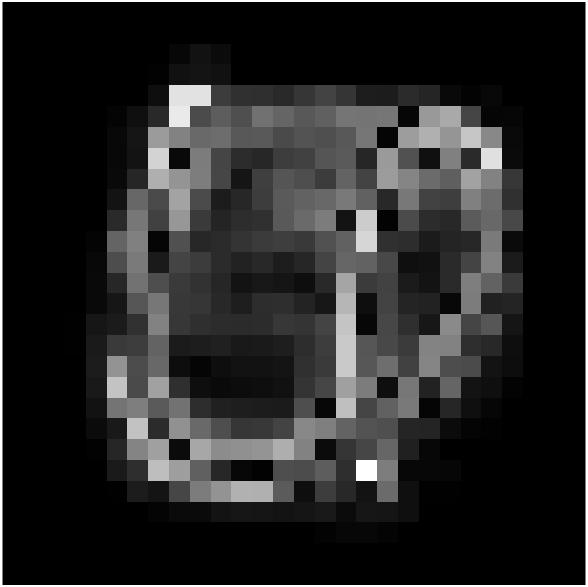}} =
    \raisebox{-.4\height}	{\includegraphics[width=.15\columnwidth]{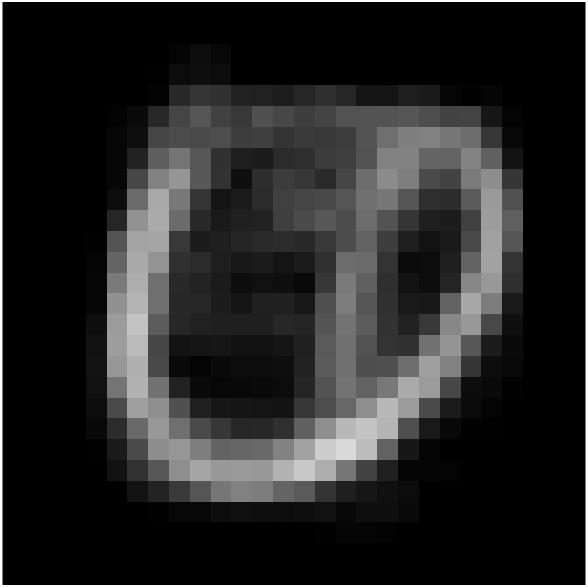}} 
    \begin{minipage}[b]{.2\columnwidth}
    {\footnotesize {(on \cvx of all classes)}}
    \end{minipage}
    \end{minipage}\\
    \begin{minipage}[b]{.48\columnwidth}
    \centering
    \raisebox{-.4\height}{\includegraphics[width=.2\columnwidth]{figures/mnist_te2119orig.jpg}} {\vspace{.2cm} -}
    \raisebox{-.4\height}	{\includegraphics[width=.2\columnwidth]{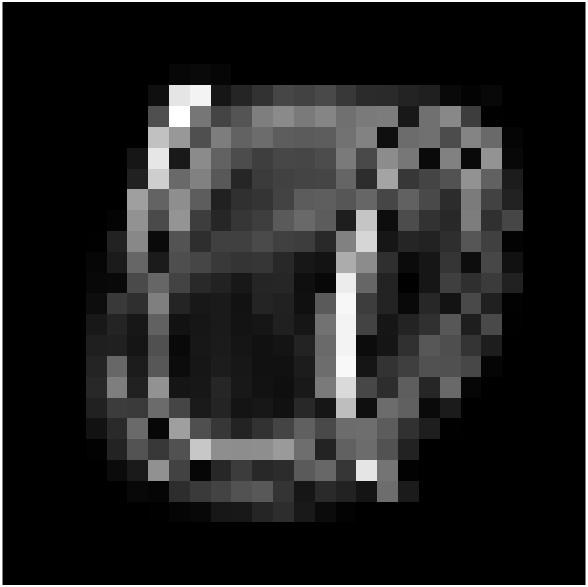}} =
    \raisebox{-.4\height}	{\includegraphics[width=.2\columnwidth]{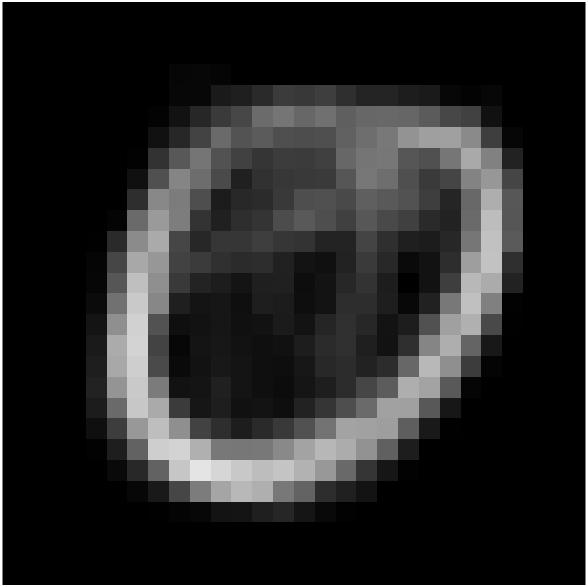}} 
    \begin{minipage}[b]{.2\columnwidth}
    {\footnotesize {(on \cvx\\ of 0s)}}
    \end{minipage}
    \end{minipage}
    \begin{minipage}[b]{.48\columnwidth}
    \centering
    \raisebox{-.4\height}{\includegraphics[width=.2\columnwidth]{figures/mnist_te2119orig.jpg}} {\vspace{.2cm} -}
    \raisebox{-.4\height}	{\includegraphics[width=.2\columnwidth]{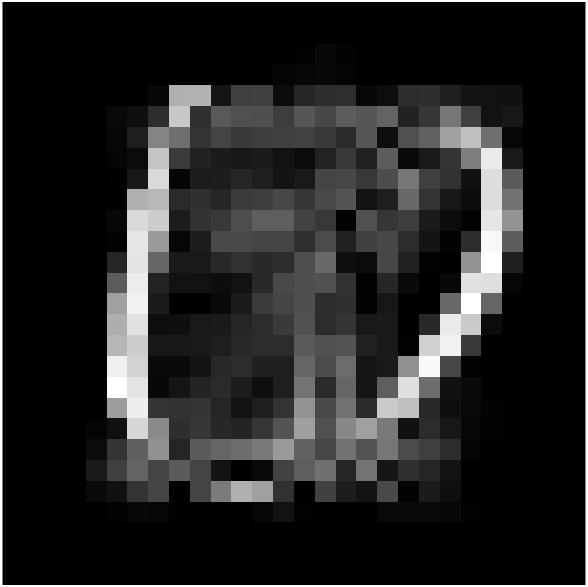}} =
    \raisebox{-.4\height}	{\includegraphics[width=.2\columnwidth]{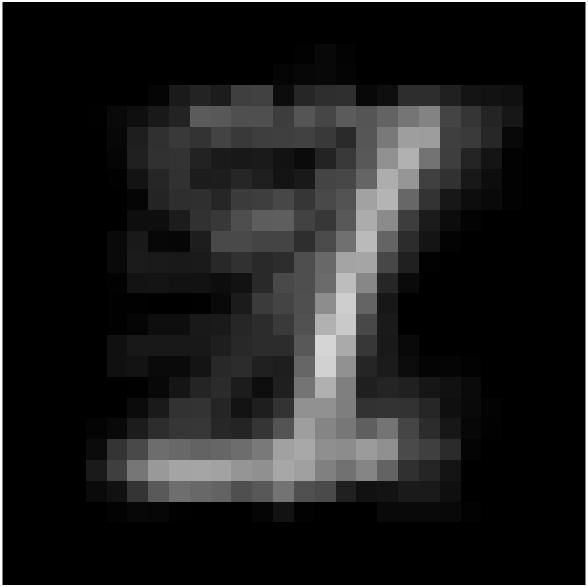}} 
    \begin{minipage}[b]{.19\columnwidth}
    {\footnotesize {(on \cvx\\ of 1s)}}
    \end{minipage}
    \end{minipage}
    \begin{minipage}[b]{.48\columnwidth}
    \centering
    \raisebox{-.4\height}{\includegraphics[width=.2\columnwidth]{figures/mnist_te2119orig.jpg}} {\vspace{.2cm} -}
    \raisebox{-.4\height}	{\includegraphics[width=.2\columnwidth]{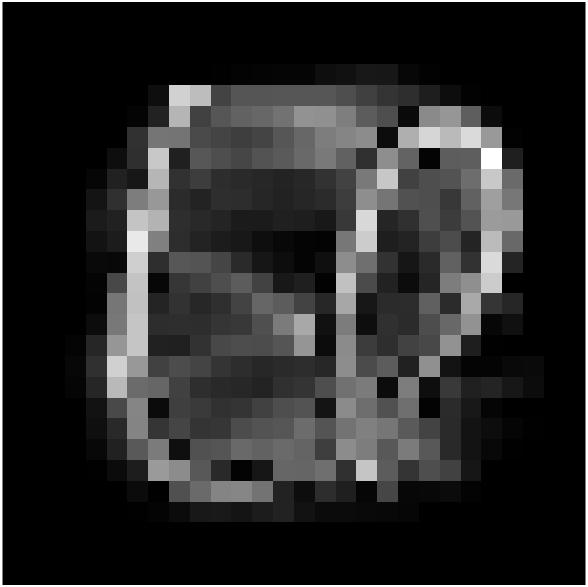}} =
    \raisebox{-.4\height}	{\includegraphics[width=.2\columnwidth]{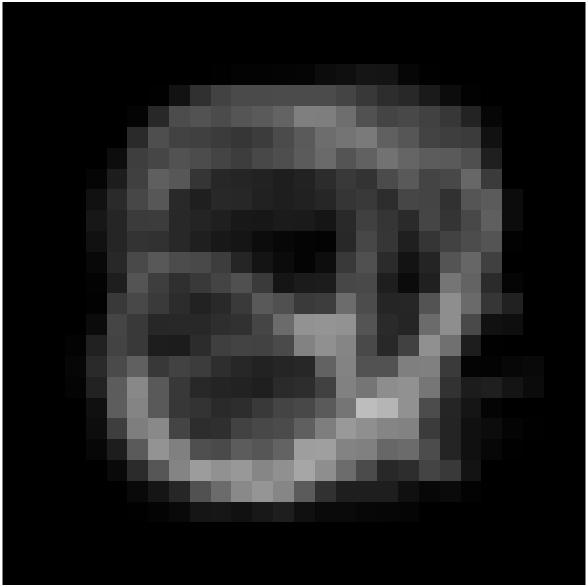}} 
    \begin{minipage}[b]{.2\columnwidth}
    {\footnotesize {(on \cvx\\ of 2s)}}
    \end{minipage}
    \end{minipage}
    \begin{minipage}[b]{.48\columnwidth}
    \centering
    \raisebox{-.4\height}{\includegraphics[width=.2\columnwidth]{figures/mnist_te2119orig.jpg}} {\vspace{.2cm} -}
    \raisebox{-.4\height}	{\includegraphics[width=.2\columnwidth]{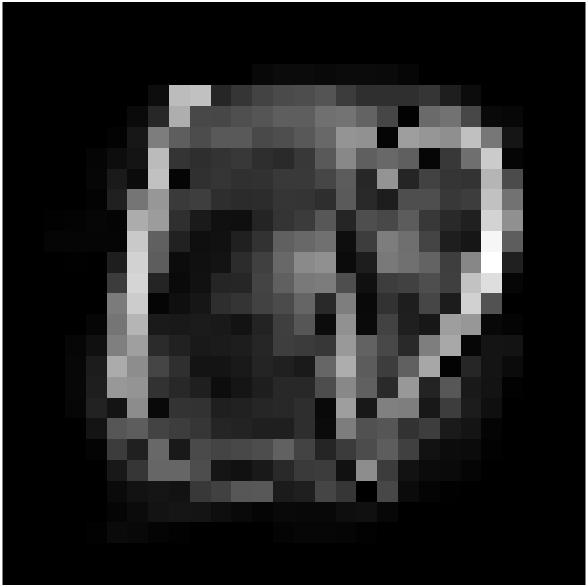}} =
    \raisebox{-.4\height}	{\includegraphics[width=.2\columnwidth]{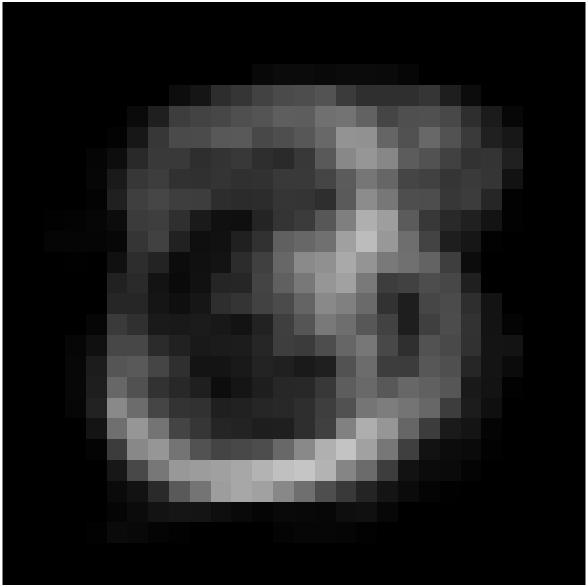}} 
    \begin{minipage}[b]{.2\columnwidth}
    {\footnotesize {(on \cvx\\ of 3s)}}
    \end{minipage}
    \end{minipage}
    \begin{minipage}[b]{.48\columnwidth}
    \centering
    \raisebox{-.4\height}{\includegraphics[width=.2\columnwidth]{figures/mnist_te2119orig.jpg}} {\vspace{.2cm} -}
    \raisebox{-.4\height}	{\includegraphics[width=.2\columnwidth]{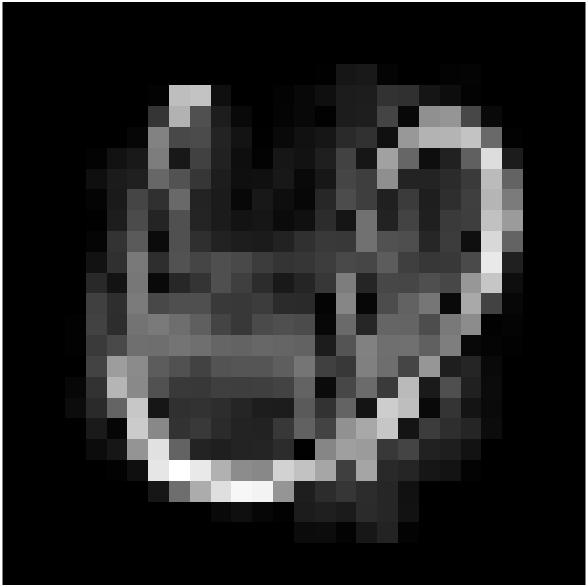}} =
    \raisebox{-.4\height}	{\includegraphics[width=.2\columnwidth]{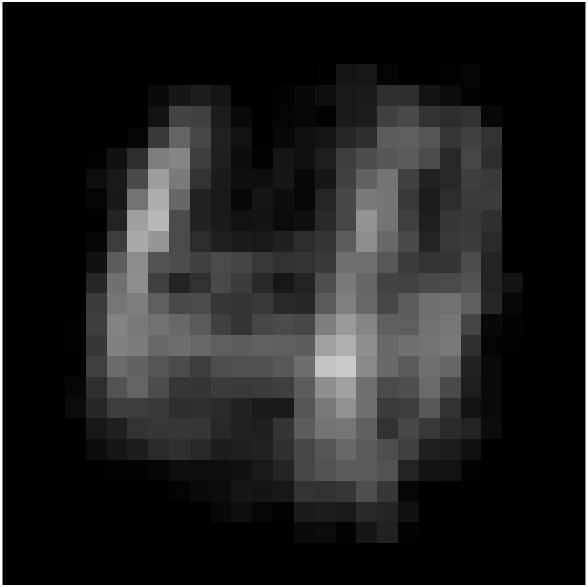}} 
    \begin{minipage}[b]{.2\columnwidth}
    {\footnotesize {(on \cvx\\ of 4s)}}
    \end{minipage}
    \end{minipage}
    \begin{minipage}[b]{.48\columnwidth}
    \centering
    \raisebox{-.4\height}{\includegraphics[width=.2\columnwidth]{figures/mnist_te2119orig.jpg}} {\vspace{.2cm} -}
    \raisebox{-.4\height}	{\includegraphics[width=.2\columnwidth]{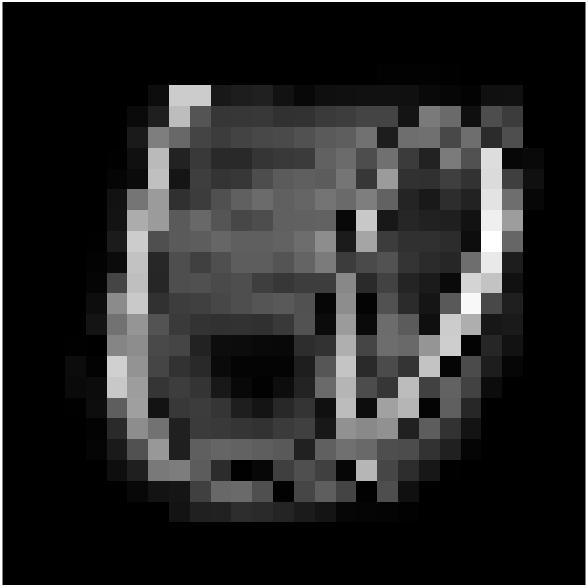}} =
    \raisebox{-.4\height}	{\includegraphics[width=.2\columnwidth]{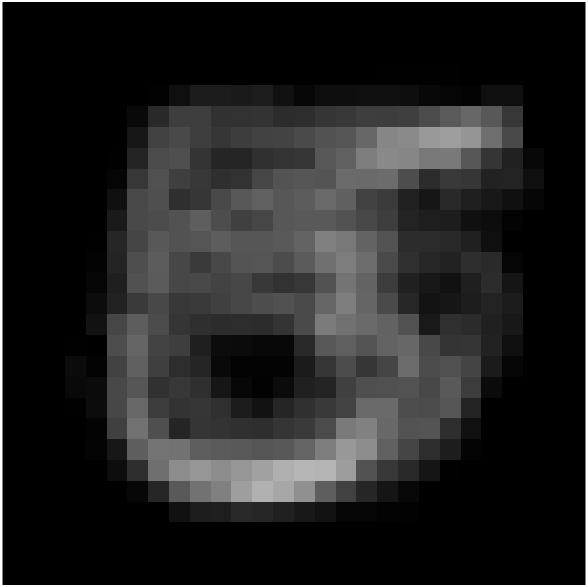}} 
    \begin{minipage}[b]{.2\columnwidth}
    {\footnotesize {(on \cvx\\ of 5s)}}
    \end{minipage}
    \end{minipage}
    \begin{minipage}[b]{.48\columnwidth}
    \centering
    \raisebox{-.4\height}{\includegraphics[width=.2\columnwidth]{figures/mnist_te2119orig.jpg}} {\vspace{.2cm} -}
    \raisebox{-.4\height}	{\includegraphics[width=.2\columnwidth]{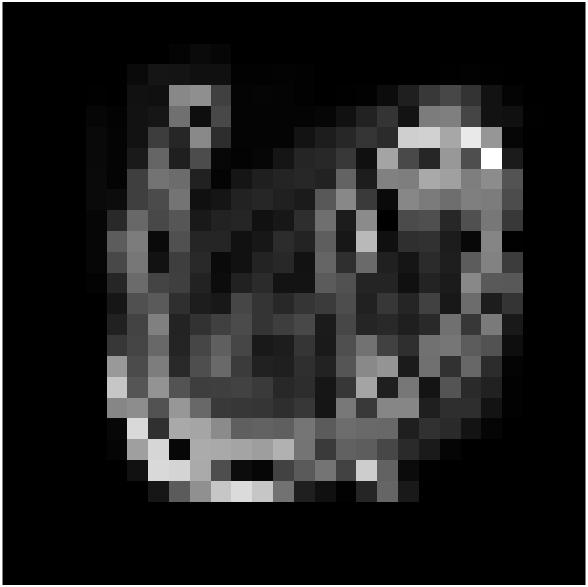}} =
    \raisebox{-.4\height}	{\includegraphics[width=.2\columnwidth]{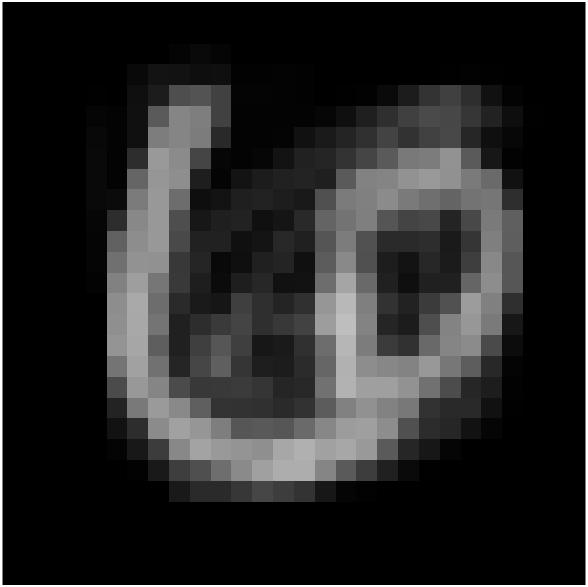}} 
    \begin{minipage}[b]{.2\columnwidth}
    {\footnotesize {(on \cvx\\ of 6s)}}
    \end{minipage}
    \end{minipage}
    \begin{minipage}[b]{.48\columnwidth}
    \centering
    \raisebox{-.4\height}{\includegraphics[width=.2\columnwidth]{figures/mnist_te2119orig.jpg}} {\vspace{.2cm} -}
    \raisebox{-.4\height}	{\includegraphics[width=.2\columnwidth]{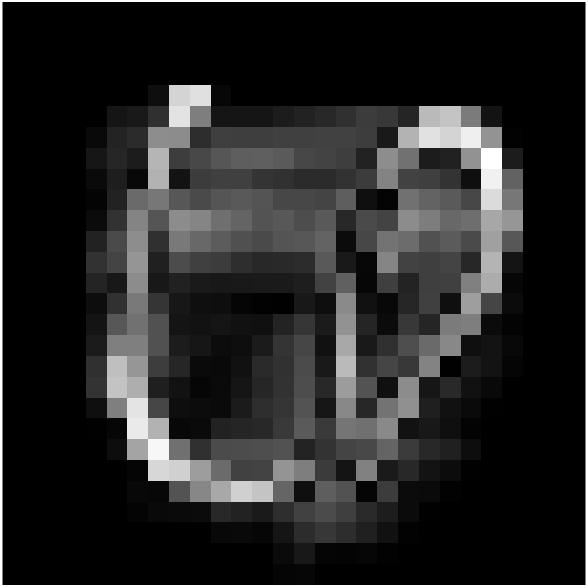}} =
    \raisebox{-.4\height}	{\includegraphics[width=.2\columnwidth]{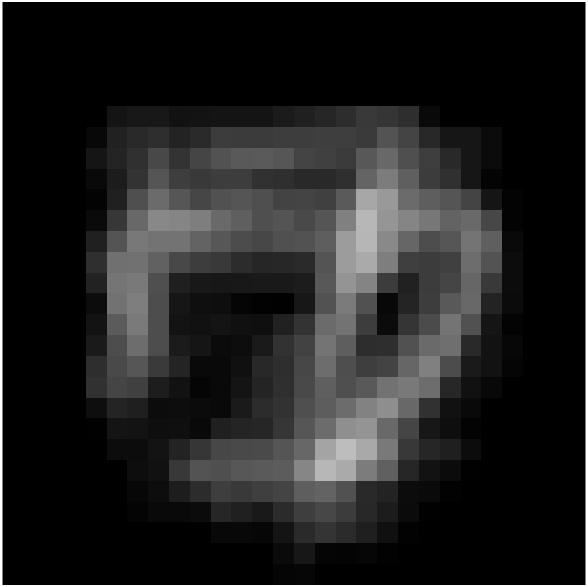}} 
    \begin{minipage}[b]{.2\columnwidth}
    {\footnotesize {(on \cvx\\ of 7s)}}
    \end{minipage}
    \end{minipage}
    \begin{minipage}[b]{.48\columnwidth}
    \centering
    \raisebox{-.4\height}{\includegraphics[width=.2\columnwidth]{figures/mnist_te2119orig.jpg}} {\vspace{.2cm} -}
    \raisebox{-.4\height}	{\includegraphics[width=.2\columnwidth]{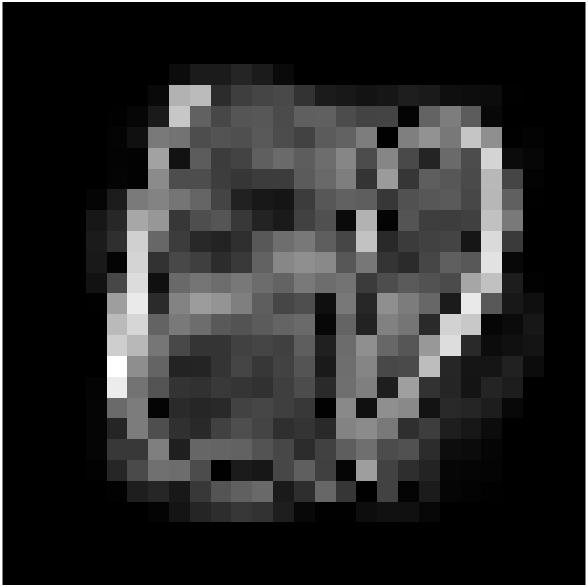}} =
    \raisebox{-.4\height}	{\includegraphics[width=.2\columnwidth]{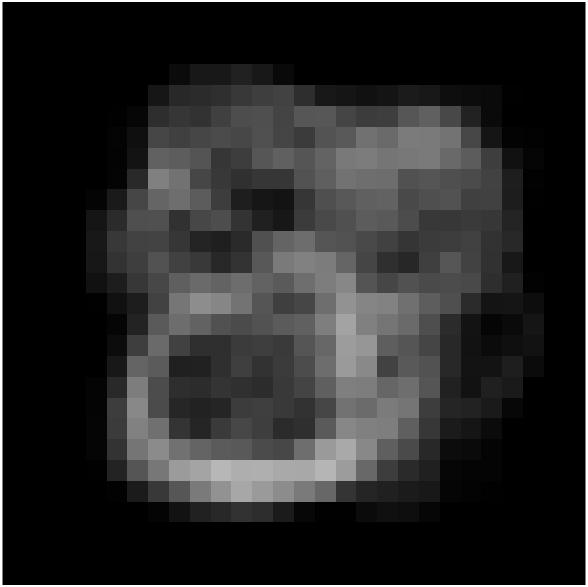}} 
    \begin{minipage}[b]{.2\columnwidth}
    {\footnotesize {(on \cvx\\ of 8s)}}
    \end{minipage}
    \end{minipage}
    \begin{minipage}[b]{.48\columnwidth}
    \centering
    \raisebox{-.4\height}{\includegraphics[width=.2\columnwidth]{figures/mnist_te2119orig.jpg}} {\vspace{.2cm} -}
    \raisebox{-.4\height}	{\includegraphics[width=.2\columnwidth]{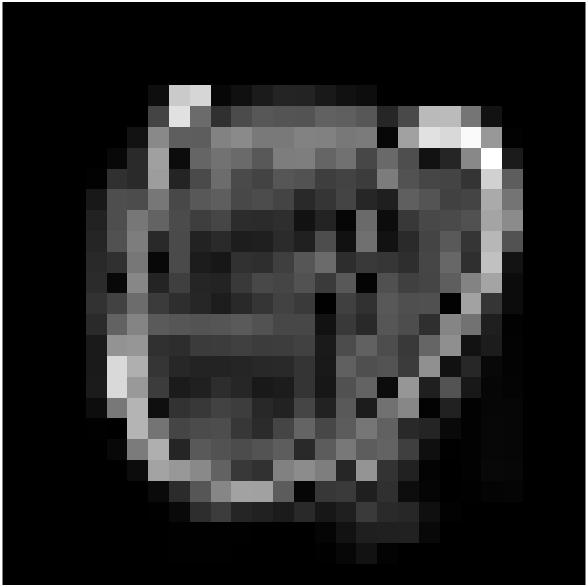}} =
    \raisebox{-.4\height}	{\includegraphics[width=.2\columnwidth]{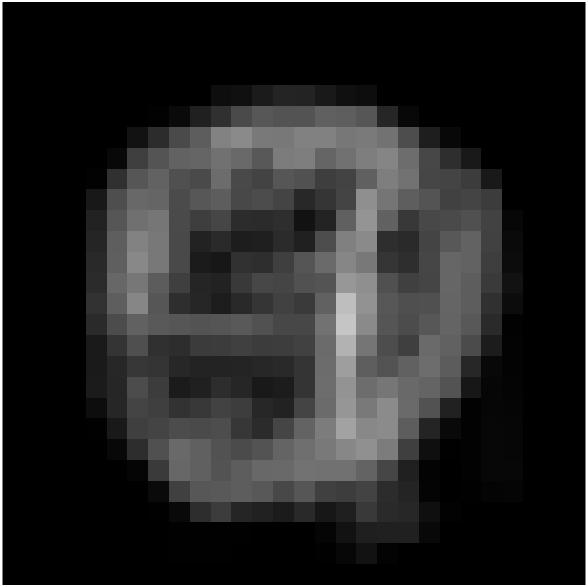}} 
    \begin{minipage}[b]{.2\columnwidth}
    {\footnotesize {(on \cvx\\ of 9s)}}
    \end{minipage}
    \end{minipage}
\end{center}
    \caption{Perturbation that can bring a testing image of MNIST to \cvx.}
  \label{fig:conv_perturb_mnist}
\end{figure}

For each image that is outside the \cvx, there is some minimum perturbation that would bring that image to the \cvx. Figures~\ref{fig:conv_perturb_cif10} and~\ref{fig:conv_perturb_mnist} show that perturbation for some images in the testing set of CIFAR-10 and MNIST.% that can bring them to their corresponding \cvx. We note that due to our approximation method, there is no guarantee that images in the middle column are exactly the minimum required perturbation, but we expect it to be sufficiently close to that minimum.

% \vspace{-.2cm}

The perturbations required to bring testing images to the \cvx specifically relate to the objects of interest depicted in images and they appear to impact the images significantly. Therefore, the extrapolation required to classify those images can be considered significant, too.

\begin{figure}[H]
\begin{center}
    \begin{minipage}[b]{.85\columnwidth}
    \centering
    (original) \raisebox{-.5\height}{\includegraphics[width=.13\columnwidth]{figures/cif10_te1orig.jpg}} {\vspace{.2cm} -}
    \raisebox{-.5\height}	{\includegraphics[width=.13\columnwidth]{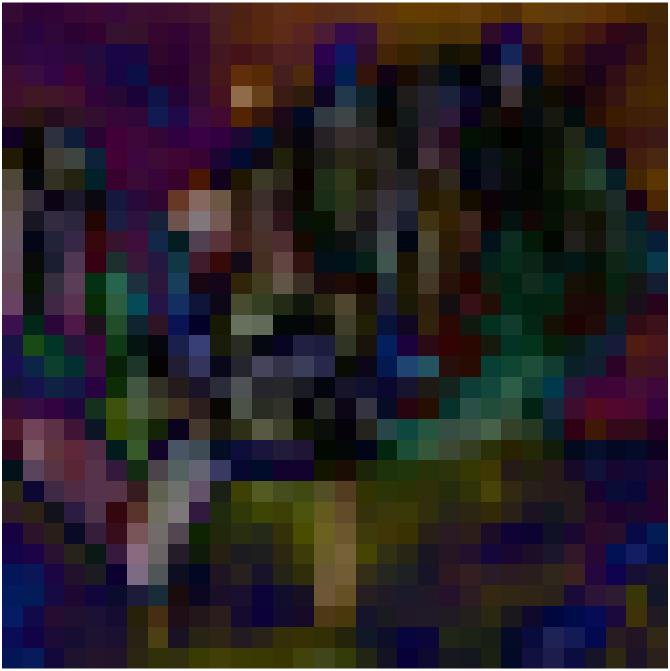}} =
    \raisebox{-.5\height}	{\includegraphics[width=.13\columnwidth]{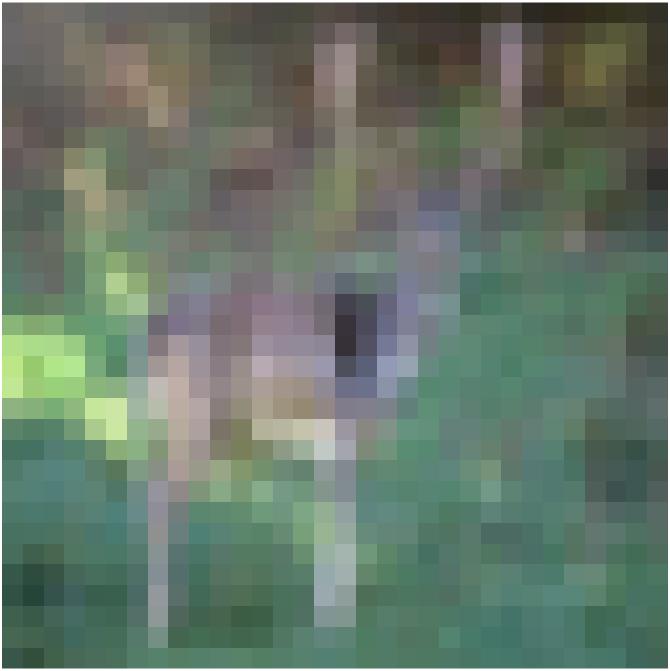}} (closest image in \cvx)
    \end{minipage}
\end{center}
    \caption{The distance between an outside image and the closest image to it in the \cvx can be considerably large, compared to the distance between the image and the boundary of \cvx. The direction between the outside image and the closest image to it is usually unrelated to the distinctive features in images.}
  \label{fig:closest_image_cif10}
\end{figure}

If we merely choose the image in training set that is closest to the query image, the distance to the query image will be considerably larger and the differences between the two images will not be related to distinctive features. Figure~\ref{fig:closest_image_cif10} shows this for the image we considered in the first row of Figure~\ref{fig:conv_perturb_cif10}, which is the very first testing image in the dataset. Hence, the interpolation performed between the training samples is useful in defining the extrapolation task.

\subsection{Images that form the surface closest to the outside image}

\begin{figure}[H]
\begin{center}
    \begin{minipage}[b]{1\columnwidth}
    \centering
    \raisebox{-.4\height}	{\includegraphics[width=.07\columnwidth]{figures/cif10_te2orig.jpg}} $\xrightarrow[]{\mathcal{H}^{tr}}$
    \raisebox{-.4\height}{\includegraphics[width=.07\columnwidth]{figures/cif10_te2oncvx_all.jpg}} {\vspace{.2cm} = 0.1129}
    \raisebox{-.4\height}	{\includegraphics[width=.07\columnwidth]{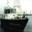}} + 0.1032
    \raisebox{-.4\height}	{\includegraphics[width=.07\columnwidth]{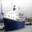}} + 0.0878
    \raisebox{-.4\height}	{\includegraphics[width=.07\columnwidth]{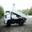}} + 0.0704
    \raisebox{-.4\height}	{\includegraphics[width=.07\columnwidth]{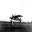}} \\ + 0.0630
    \raisebox{-.4\height}	{\includegraphics[width=.07\columnwidth]{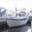}} + 0.0536
    \raisebox{-.4\height}	{\includegraphics[width=.07\columnwidth]{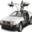}} + 0.0533
    \raisebox{-.4\height}	{\includegraphics[width=.07\columnwidth]{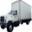}} + 0.0529
    \raisebox{-.4\height}	{\includegraphics[width=.07\columnwidth]{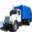}} + 0.0516
    \raisebox{-.4\height}	{\includegraphics[width=.07\columnwidth]{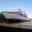}} \\ + 0.0427
    \raisebox{-.4\height}	{\includegraphics[width=.07\columnwidth]{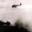}} + $\dots$
    \end{minipage}
    \end{center}
    \caption{Projection of a testing image to the \cvx, and the convex combination of images in training set that create the projected image. There are 26 training images used in the convex combination, but we have shown 10 of those with the largest coefficients.}
  \label{fig:cvx_support}
\end{figure}

The image on the surface of \cvx, closest to an image, $x^{te}$, outside the \cvx, is a convex combination of some training images. We call those training images as the support for $x^{te}$. In our experiments, the number of support images for each testing image is only a few. For example, the image on the second row of Figure~\ref{fig:conv_perturb_cif10} has only 26 support images in the training set. Figure~\ref{fig:cvx_support} shows 10 of those 26 images with their corresponding coefficients in the convex combination.

% \vspace{-.2cm}

\subsection{Internal representation of images learned by deep networks}

The success of deep learning models is attributed to the features they extract from images before their final classification layer. Let's consider a standard ResNet model \citep{he2016deep} trained on the CIFAR-10 training set. The internal layer of network just before the last pooling layer has 4,096 dimensions. In that space, testing samples are outside the convex hull of training set, similar to the pixel space. Hence, all the operations and feature extractions performed on images, from pixel space all the way to the last pooling layer do not bring the testing samples inside the convex hull of training sets.

\begin{figure}[H]
    \centering
    \includegraphics[width=.5\linewidth]{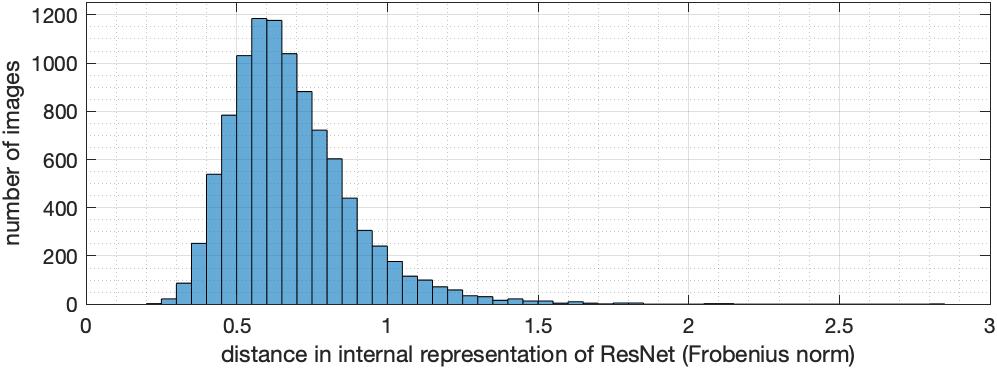}
    \caption{Variations of distance to $\mathcal{H}^{tr}$ in the 64-dimensional representations of images after the last pooling layer of a trained ResNet model (CIFAR-10). All testing images are outside the $\mathcal{H}^{tr}$.}
  \label{fig:conv_dist_internal_pool}
\end{figure}

The last pooling layer then projects the images to a 64 dimensional space, and consequently to a 10-dimensional space which corresponds to the 10 output classes of network. Even in the 64-dimensional space, all testing images are outside the \cvx while their distances resemble a normal distribution as shown in Figure~\ref{fig:conv_dist_internal_pool}.% Remember that in the pixel and wavelet space, almost all testing samples were outside the \cvx and their distances resembled a normal distribution as we showed in Figure~\ref{fig:conv_dist}.

% Figure~\ref{fig:pooling_out_samples} shows the first 5 testing images that are outside the convex hull of training set in that 64-dimensional representations of images, and

\begin{figure}[H]
    \centering
         \includegraphics[width=.1\linewidth]{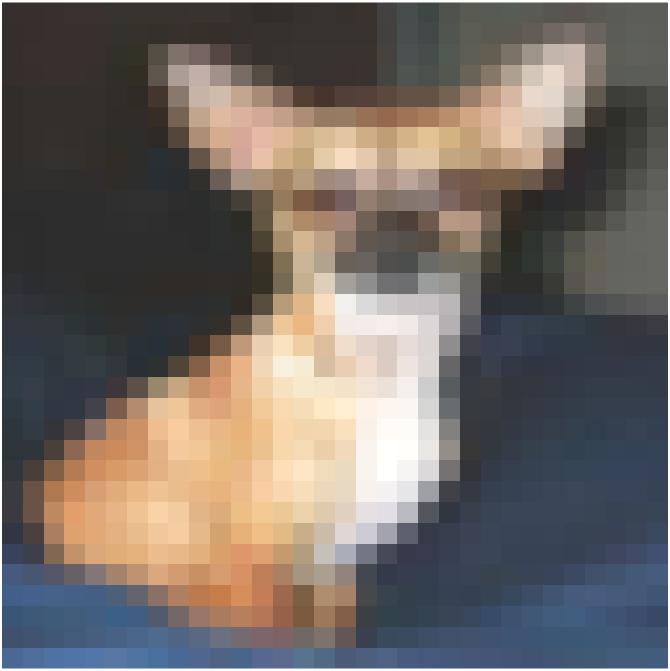}
         \includegraphics[width=.1\linewidth]{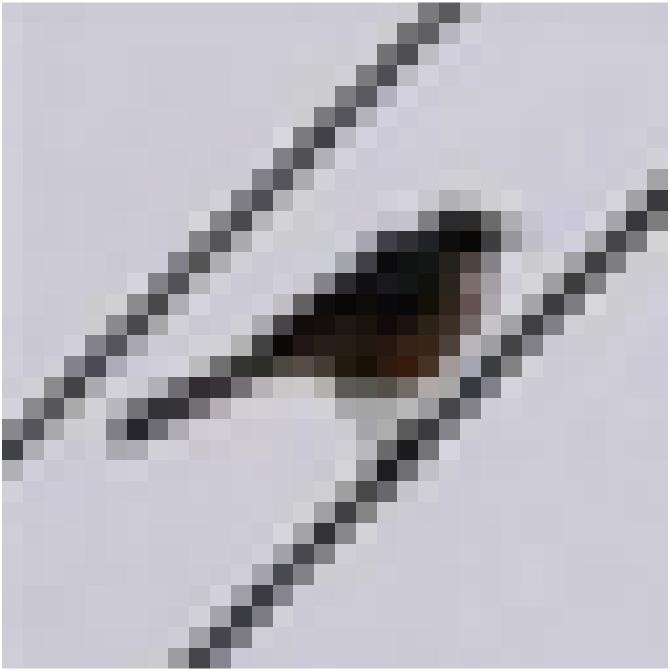}
         \includegraphics[width=.1\linewidth]{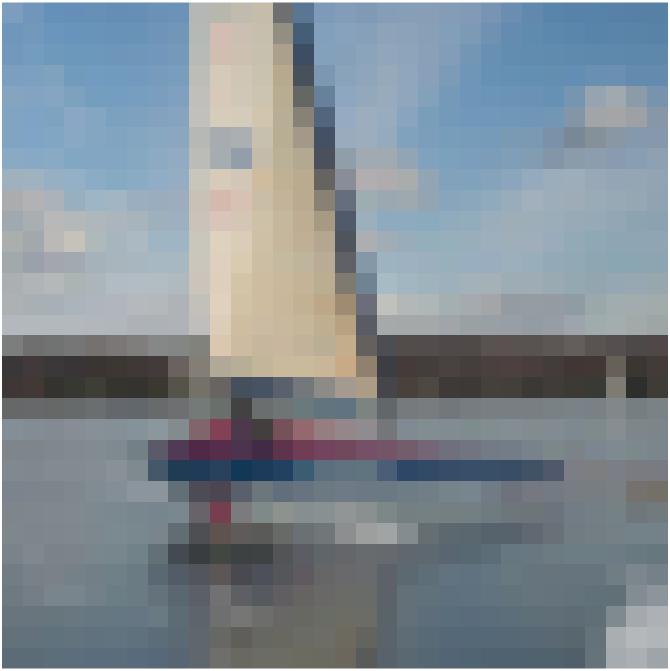}
         \includegraphics[width=.1\linewidth]{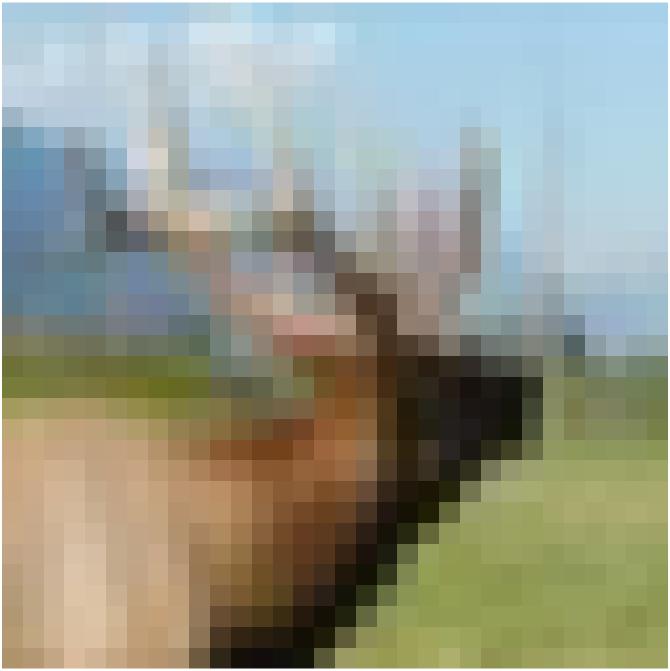}
         \includegraphics[width=.1\linewidth]{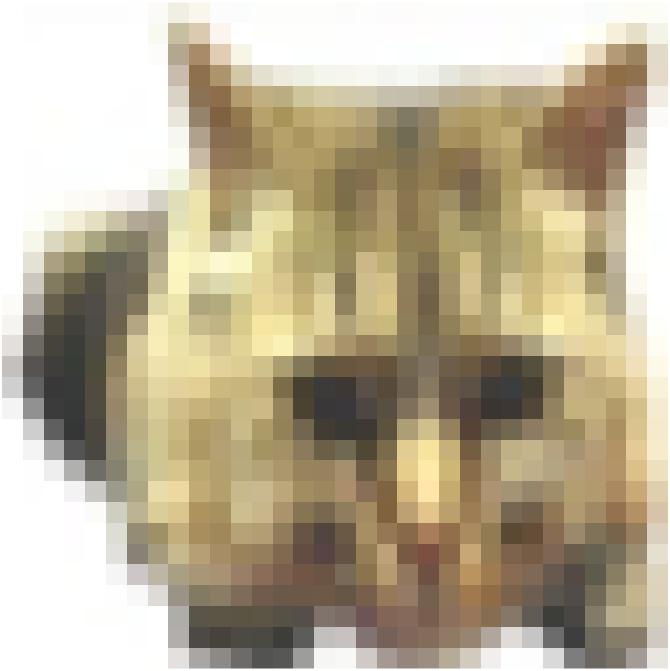}
    \caption{The 5 testing images that are farthest from the \cvx in the internal 64-dimensional space after the last pooling layer of a trained ResNet model. Compare these to farthest images from \cvx in the pixel space as shown in Figure~\ref{fig:far_cif10}.}
  \label{fig:pooling_far_samples}
\end{figure}

Figure~\ref{fig:pooling_far_samples} shows the testing images that are farthest from \cvx in that 64-dimensional representations of images.% For some of these testing images, there are considerably similar images in the training set.

% \begin{figure}[H]
%     \centering
%          \includegraphics[width=.12\linewidth]{figures/cif10_te3pooling1.jpg}
%          \includegraphics[width=.12\linewidth]{figures/cif10_te4pooling2.jpg}
%          \includegraphics[width=.12\linewidth]{figures/cif10_te51pooling3.jpg}
%          \includegraphics[width=.12\linewidth]{figures/cif10_te56pooling4.jpg}
%          \includegraphics[width=.12\linewidth]{figures/cif10_te64pooling5.jpg}
%     \caption{The first 5 testing images that are outside the convex hull of training set in the internal 64-dimensional space after the last pooling layer of a trained ResNet model.}
%   \label{fig:pooling_out_samples}
% \end{figure}

It may be speculated that there is a low dimensional manifold, learned by deep networks, where all (or most) testing samples are included in the convex hull of training set and the functional task of a deep network on that manifold is interpolation. We observed here that there is no trace of such manifold even in the internal representation of a trained ResNet. We saw that even in the 64-dimensional layer right before the classification layer, testing samples are outside the \cvx and their distance to \cvx is still significant.

In fact, in all the analysis presented we found no indication that the task of classifying testing images of our datasets can be considered an interpolation task.

%It may be speculated that there is a 64-dimensional manifold where most testing images are inside the \cvx. %But, via the adversarial attacks literature, we well know how easy it is to move across classes in that manifold by making imperceivable changes in the pixel space. Imperceivable changes in the pixel space can move the testing images outside the \cvx, too. 
%While such manifold is not intuitive, it is important to realize that we only need 650 parameters to classify the output of 64-dimensional layer. The 650 parameter portion of the model is only a tiny fraction of the model and not an over-parameterized system. The huge number of parameters in the network are involved in projecting the images from the pixel space into that internal 64-dimensional space. Hence, the main extrapolation task of the model that requires over-parameterization is performed in layers prior to the 64-dimensional layer.

%Via the adversarial attacks literature, we well know how easy it is to move across classes in that manifold by making imperceivable changes in the pixel space. Imperceivable changes in the pixel space can move the testing images outside the \cvx, too. 

% These observations reiterate the extrapolation nature of classifying unseen images.

\subsection{Takeaways from geometric properties of datasets}

We can conclude that with such number of training samples in such high-dimensional domains, one can expect the testing samples to be outside their \cvx. However, directions to the convex hulls are significant in size, and they contain information related to the objects of interest in images. By investigating the representation of images in internal layers of a standard ResNet model, we observed that testing images are outside the convex hull of training set, all along the layers of the network, even after the last pooling layer which only has 64-dimensions. All our analysis indicated that classifying images of these datasets is generally an extrapolation task (to a moderate degree).

Therefore, the generalization of over-parameterized image classifiers cannot be explained by mere interpolation. Instead, we have to consider interpolation in tandem with extrapolation. Interpolation defines the decision boundaries inside the \cvx and brings them to the surface of \cvx. Extrapolation helps us shape their extensions to some degree outside the \cvx.

When a human (even a young child) classifies the object in an image (e.g., a car), it is hard to imagine that they are interpolating between images of that object available in their memory. Even if we consider a very large training set of images of an object, it would be relatively easy to come up with a legitimate image of that object that falls outside the convex hull of that large training set.

It may be argued that location of decision boundaries are not important in the pixel space, but that would be ignoring the domain of our function. How can the domain not be important? Previous studies such as \citep{elsayed2018large,marginbased2019} have shown that pushing the decision boundaries away from the training samples in the pixel space improves the generalization of models. Adversarial examples are also created in the pixel space and another line of research is trying to push the decision boundaries away from samples to make the models less vulnerable \citep{cohen2019certified}. Therefore location of decision boundaries in the domain is actually what researchers have been using to understand and change the main functional traits of our deep networks. In fact, even basic training techniques such as mirroring training images (horizontally) and feeding them to the model as extra training samples, aim to define decision boundaries in more regions of the domain.

This is not to say that if we train a model to classify images of shoes, it can go on to extrapolate outside the scope of what it has learned, and classify the contents of any image, e.g., classify tumors in radiology images of liver. We do not use the ``extrapolation" in that sense. When we studied the \cvx of random points (in Appendix~\ref{appx:random_points}), we clearly saw that there is a close relationship between the training and testing sets of our image datasets, and we also know from the literature that there is an intimate relationship between the spatial location of decision boundaries of the models and their training and testing sets. Nevertheless, it is mathematically proper to say that the task of classifying unseen images can be an extrapolation task, outside the \cvx of what the model has seen/learned to some moderate degree.

% \footnote{For example, mirroring the contents of training images (horizontally) and using them as extra training samples, are well-known techniques that help with generalization of models. We can interpret such technique as defining decision boundaries in regions of the 

To this end, we can consider 2 levels of extrapolation: 1- Extrapolating moderately outside the training set, but within the scope of training set, (moderate may be some fraction of the diameter of \cvx, for example) and 2- Extrapolating outside that scope. We use the term ``extrapolation" in the former sense and not the latter.

The latter case relates to the out-of-distribution detection literature in deep learning. Understanding the distribution of image datasets has been a challenging task and we still do not have reliable tools to detect out-of-distribution images. For example, a model trained on CIFAR-10 may confidently classify an MNIST digit as a cat without even realizing that a handwritten digit has no similarity to what it has seen during training. Numerous papers have tried to address this shortcoming of deep networks, but it is still an open research problem. We believe considering the geometry of data may be useful in the out-of-distribution detection, too.

We have to remain mindful that in the pixel space, spatial locations and Euclidean distances are not always meaningful. For example, we can move considerable distances in the pixel space by changing the background of an image of a cat, or by changing the color of the cat, or by moving the cat from one location in the image to another location. Therefore, drawing an exact line as to which regions outside the \cvx should be considered out-of-distribution is not so easy. This also relates to recent empirical studies on the foreground/background effect of images on deep learning generalization \citep{xiao2020noise}. We believe the geometric perspective we propose in this paper can be a useful guide for such studies.

The perspective we propose here remains consistent in all these subfields of the literature. When we train a model, we define its decision boundaries inside its \cvx, and those decision boundaries extend outside the \cvx all over the domain, even to areas where we do not have any training samples. By over-parameterization and training regimes, we manage to form the extensions of our decision boundaries to some extent outside the \cvx and that is our key to classifying testing images. Nevertheless, we do not have control over how decision boundaries extend to all regions outside the \cvx. For training a network, we gradually minimize the loss function of the model on training set, while observing how that minimization affects the accuracy of model on a validation set. Watching the validation set is our guide to shaping the extensions of decision boundaries while we train a model.

%Some areas far outside the \cvx should be considered out-of-distribution, and the model should abstain from classifying images in those regions.

Our definition of extrapolation corresponds to the definition of generalization. In the literature, when one says a deep network generalizes to classify unseen images, they do not mean that it generalizes from classifying handwritten digits to classifying tumors in radiology images. They mean that the model generalizes to classify unseen images related to images that it has been trained on. We use the term extrapolation in that same sense.

For our purpose of studying generalization, we are concerned with how the extensions of decision boundaries are formed to some moderate degree outside its \cvx, and how it relates to over-parameterization, training regime, and accuracy of model on testing sets. We will explore them in the next section.

\section{Learning outside the convex hull: A polynomial decision boundary} \label{sec:polynomial}

In the previous section, we showed that all testing samples of MNIST and CIFAR-10 are considerably outside the convex hull of their corresponding training sets, while the distance to the $\mathcal{H}^{tr}$ has noticeable variations, resembling a normal distribution. Hence, a trained deep learning model somehow manages to define its decision boundaries accurately enough outside the boundaries of what it has observed during training. But how does a model achieve that, or more precisely, how do we manage to train a model such that its decision boundaries have the desirable form outside the $\mathcal{H}^{tr}$?

% Inside the $\mathcal{H}^{tr}$, the location of decision boundaries are to be expected, because the model has s

Since we are interested in the generalization of image classifiers, and the pixel space is a bounded space, we consider the domain to be bounded, while the $\mathcal{H}^{tr}$ occupies some portion of it. Testing data can be inside and outside the $\mathcal{H}^{tr}$, but always inside the bounded domain.

Let's now use a polynomial decision boundary as an example to gain some intuitive insights.\footnote{Choosing a polynomial decision boundary is appropriate because many recent studies on deep learning generalization consider regression models that interpolate, e.g. \citep{belkin2018understand,belkin2018overfitting,belkin2019reconciling,liang2020just,verma2019manifold,kileel2019expressive,savarese2019infinite}. However, we note that those studies do not consider the convex hull of training sets.} Figure~\ref{fig:data} shows two point sets colored in blue and red, each set belonging to a class. These sets are non-linearly separable, because they have no overlap. If we use the polynomial 
\begin{equation} \label{eq:poly}
    y = 10^{-5} (x+20)(x+17)(x+10)(x+5)(x)(x-2)(x-9),
\end{equation}
as our decision boundary, we achieve perfect accuracy in separating these two sets, as shown in Figure~\ref{fig:polynomial}.

\begin{figure}[H]
\centering
     \begin{subfigure}[b]{0.35\textwidth}
         \centering
         \includegraphics[width=1\linewidth]{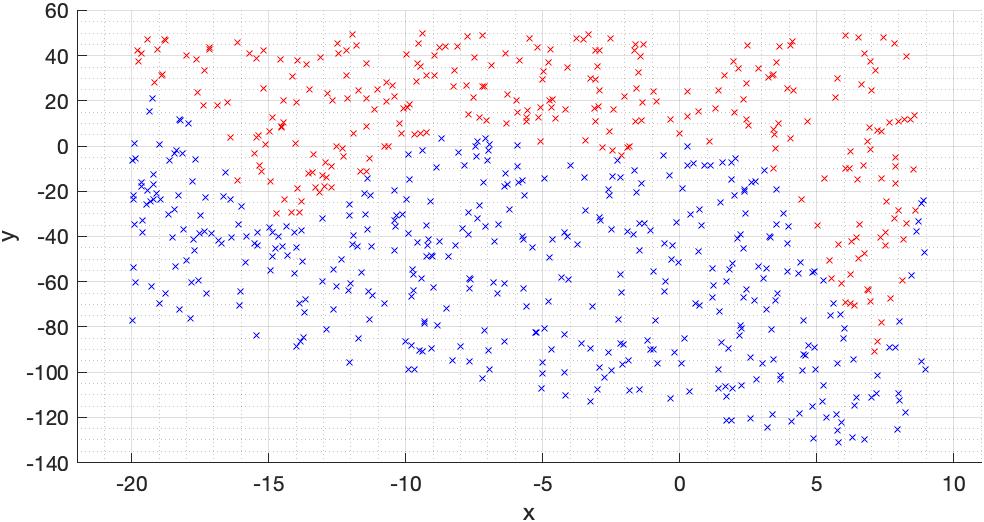}
         \caption{}
         \label{fig:data}
     \end{subfigure}
     \quad
     \begin{subfigure}[b]{0.35\textwidth}
         \centering
         \includegraphics[width=1\linewidth]{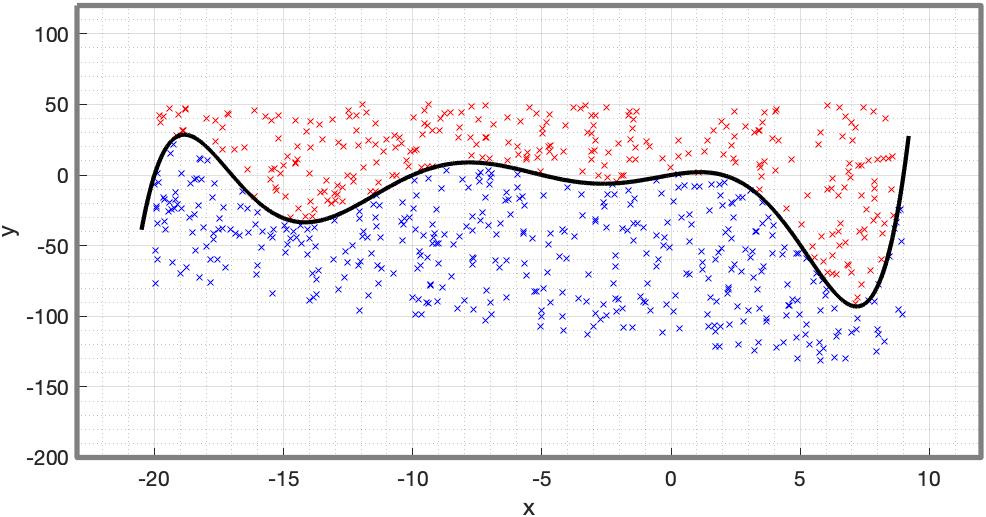}
         \caption{}
         \label{fig:polynomial}
     \end{subfigure}
  \caption{\textbf{(a)} Training data with 2 classes, colored with blue and red. \textbf{(b)} Non-linear separation of 2 classes with a polynomial of degree 7.}
  \label{fig:data_polynomial}
\end{figure}

\begin{figure}[H]
  \centering
   \includegraphics[width=0.4\linewidth]{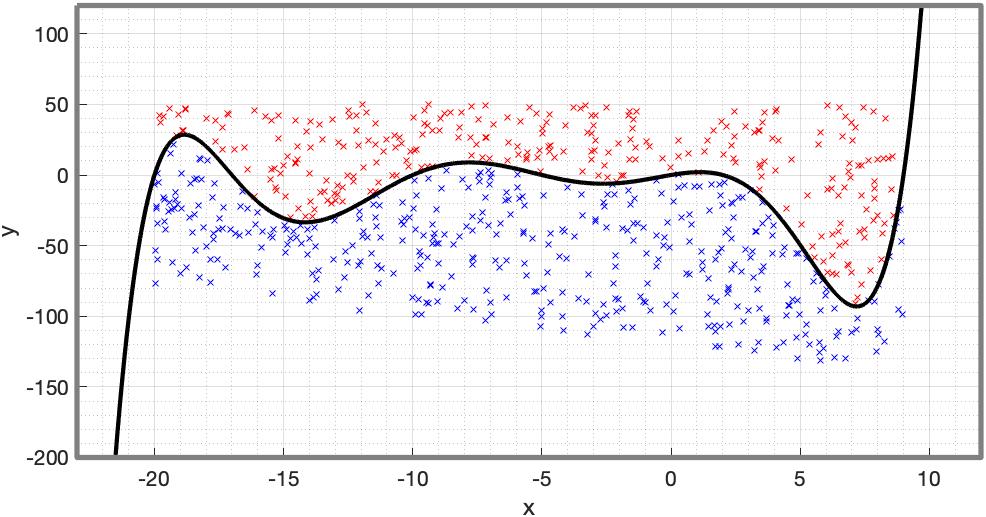}
   \caption{Shape of the polynomial decision boundary in our bounded domain, inside and outside the convex hull of its training data.}
  \label{fig:polynomial_extension}
\end{figure}

Now that we have obtained this polynomial, i.e., decision boundary, we would be interested to know how it generalizes to unseen data. Let's assume that our bounded domain is defined by the limits shown in Figure~\ref{fig:polynomial_extension} which also shows how our decision boundary generalizes outside the $\mathcal{H}^{tr}$. If our polynomial can correctly separate and label our testing data, we would say that our polynomial is generalizing well, and vice versa. But, what is reasonable to expect from the testing data? In what regions of the domain should we be hopeful that our polynomial can generalize? What if the domain is much larger than the $\mathcal{H}^{tr}$? Is the extension of our polynomial on both sides reasonable enough?

Clearly, the answer to the above questions can be different inside and outside the $\mathcal{H}^{tr}$. Inside the $\mathcal{H}^{tr}$, if the unseen data has a similar label distribution as the training set, we can be hopeful that our decision boundary will generalize well. However, outside the $\mathcal{H}^{tr}$ is uncharted territory and hence, there will be less hope/confidence about the generalization of our decision boundary, especially when we go far outside the $\mathcal{H}^{tr}$.

\begin{figure}[H]
\centering
     \begin{subfigure}[b]{0.4\textwidth}
        \centering
        \includegraphics[width=1\linewidth]{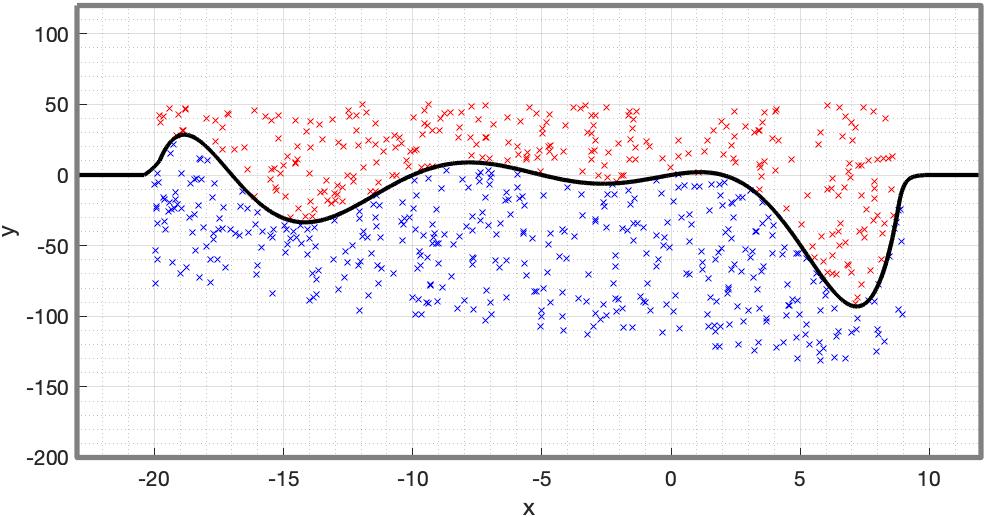}
        \caption{Possible extension of polynomial decision boundary in the over-parameterized regime.}
        \label{fig:data}
     \end{subfigure}
     \quad
     \begin{subfigure}[b]{0.4\textwidth}
        \centering
        \includegraphics[width=1\linewidth]{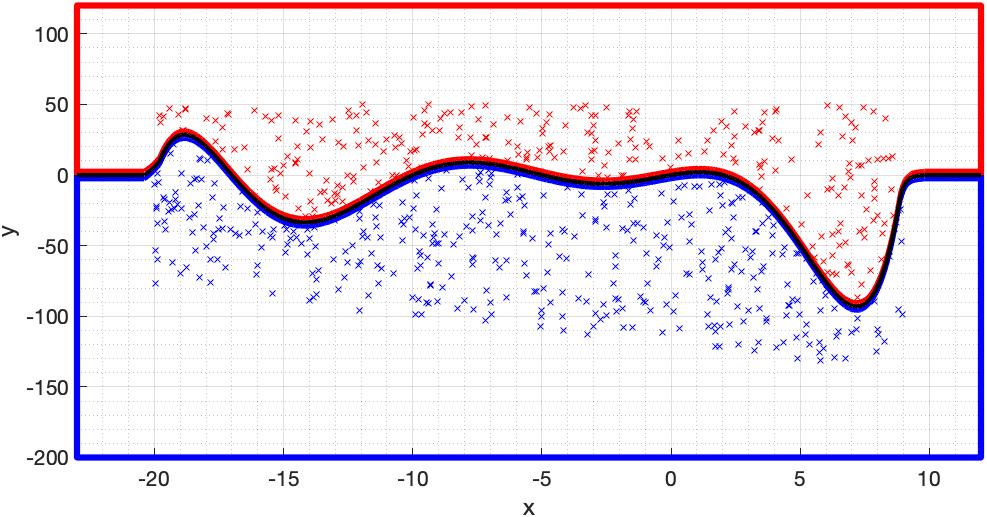}
        \caption{Resulting division of domain between the two classes, defined by red and blue bounds.}
        \label{fig:polynomial}
     \end{subfigure}
   \caption{Consider the decision boundary depicted in \textbf{(a)} and assume that the distribution of testing data is such that the red and blue bounded regions in \textbf{(b)} are densely filled with red and blue data points, respectively. It follows that the decision boundary in Figure~\ref{fig:polynomial_extension} generalizes poorly for testing points outside the $\mathcal{H}^{tr}$, despite the fact that it perfectly fits the training data.}
  \label{fig:polynomial_extension_flat}
\end{figure}

Now, let's assume that from some prior knowledge, we know that the decision boundary in Figure~\ref{fig:polynomial_extension_flat} is the unique decision boundary that perfectly classifies the testing data. In such case, the decision boundary defined by equation~\eqref{eq:poly} and shown in Figure~\ref{fig:polynomial_extension} will generalize poorly outside the $\mathcal{H}^{tr}$, despite the fact that it perfectly fits the training data.% At the same time, its generalization could be fairly acceptable at regions of the domain that are very close to the $\mathcal{H}^{tr}$.

How can we incorporate that prior knowledge into the decision boundary defined by equation~\eqref{eq:poly} and reshape it to the decision boundary in Figure~\ref{fig:polynomial_extension_flat}, so that it can generalize well both inside and outside the $\mathcal{H}^{tr}$? How can we change the shape of our polynomial outside the $\mathcal{H}^{tr}$, while maintaining its current shape inside the $\mathcal{H}^{tr}$? Clearly, we should add to the degree of our polynomial, or in other words, we should \textbf{over-parameterize} it. The necessity of over-parameterization for achieving that goal for our polynomial decision boundary can be rigorously shown using the orthogonal system of Legendre polynomials \citep{ascher2011first}. 

From this perspective, over-parameterization is necessary, but it is not sufficient for good generalization, because for an over-parameterized polynomial (i.e., decision boundary), there will be infinite number of solutions that can fit the training data, but each of them would have a different shape outside the $\mathcal{H}^{tr}$. In fact, an over-parameterized polynomial can have the same shape as the polynomial in Figure~\ref{fig:polynomial_extension}. But, how can we pick the decision boundary that fits the data and also generalizes well outside the $\mathcal{H}^{tr}$?

In the over-parameterized regime, the key to finding the desirable decision boundary is the optimization process, i.e., \textbf{the training regime}. In other words, different training regimes would lead us to decision boundaries that all perfectly fit the training set, but each has a different shape outside the $\mathcal{H}^{tr}$. This highlights that over-parameterization and training regime work in tandem to shape the extensions of our decision boundary. However, many studies consider them separately in the deep learning literature.

\section{Output of deep learning functions outside their $\mathcal{H}^{tr}$} \label{sec:deep_output}

\begin{figure}[H]
  \centering
     \begin{subfigure}[b]{0.3\textwidth}
         \centering
         \includegraphics[width=1\linewidth]{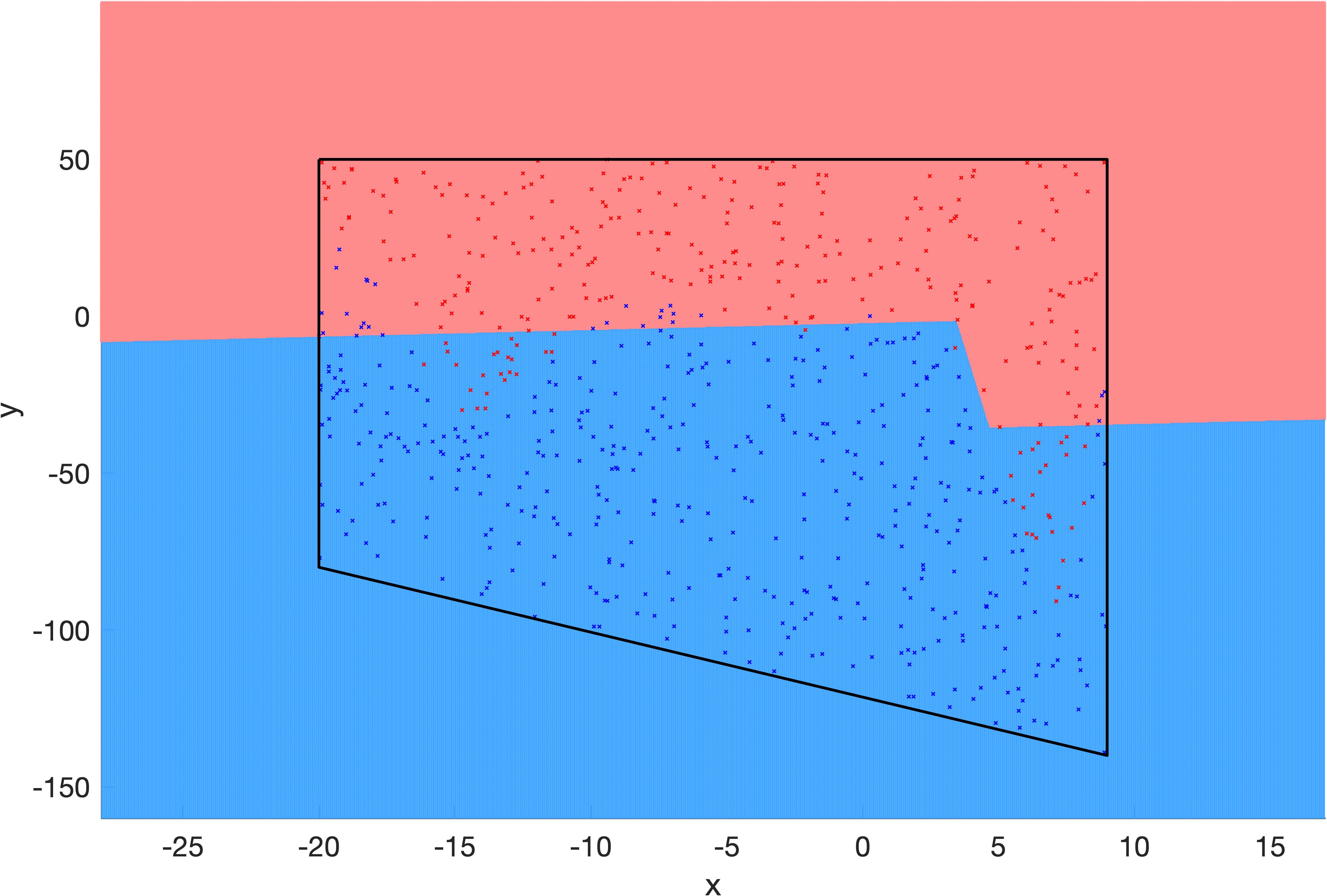}
         \caption{model with 2 neurons}
         \label{fig:nn2}
     \end{subfigure}
    %  \quad
     \begin{subfigure}[b]{0.3\textwidth}
         \centering
         \includegraphics[width=1\linewidth]{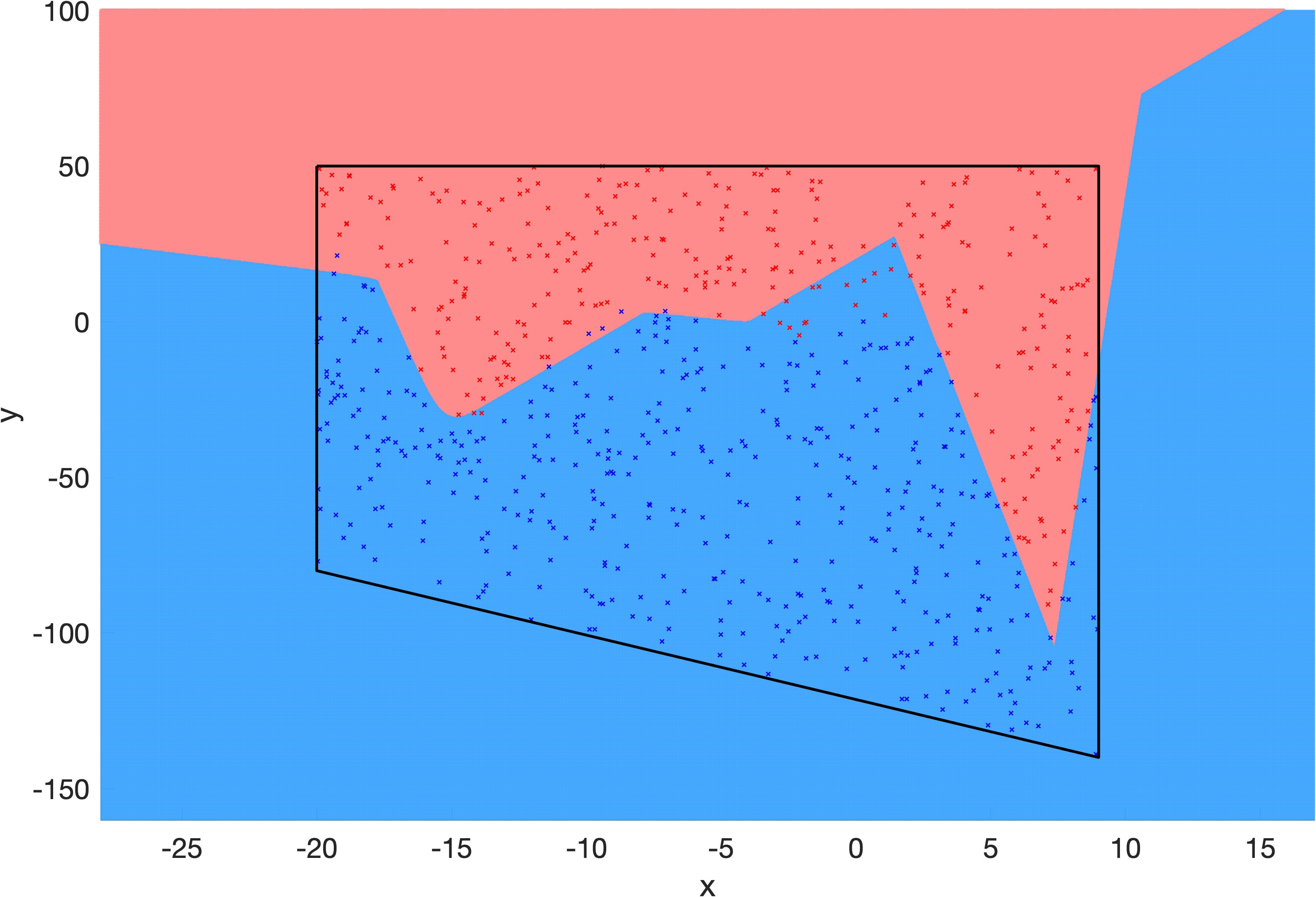}
         \caption{model with 5 neurons}
         \label{fig:nn5}
     \end{subfigure}
    %  \quad
     \begin{subfigure}[b]{0.3\textwidth}
         \centering
         \includegraphics[width=1\linewidth]{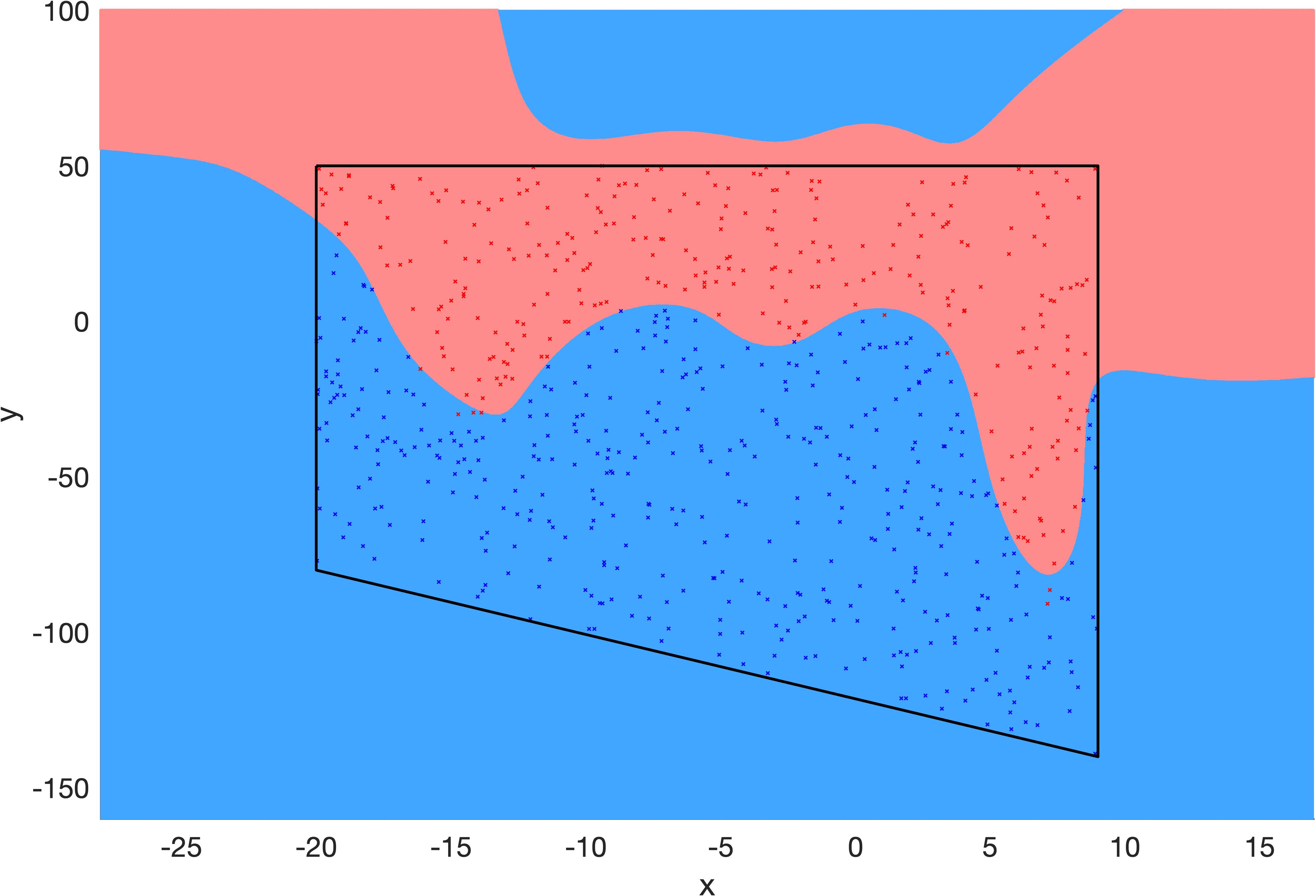}
         \caption{model with 10 neurons}
         \label{fig:nn10}
     \end{subfigure}
     \caption{Variations of model output inside and outside the $\mathcal{H}^{tr}$, for under-parameterized models. Feed-forward ReLU networks with 2,5, and 10 neurons do not have enough capacity to perfectly fit the training set. As we increase the number of neurons from 2 to 10, the model fits the training data better, while it starts to have more variations outside the \cvx. In this under-parameterized regime, we cannot minimize the training loss to zero, but each time that we train a model, we can achieve the same non-zero training loss for it, leading to the same model.}
  \label{fig:model_output_under}
\end{figure}

\begin{figure}[h]
  \centering
     \begin{subfigure}[b]{0.3\textwidth}
         \centering
         \includegraphics[width=1\linewidth]{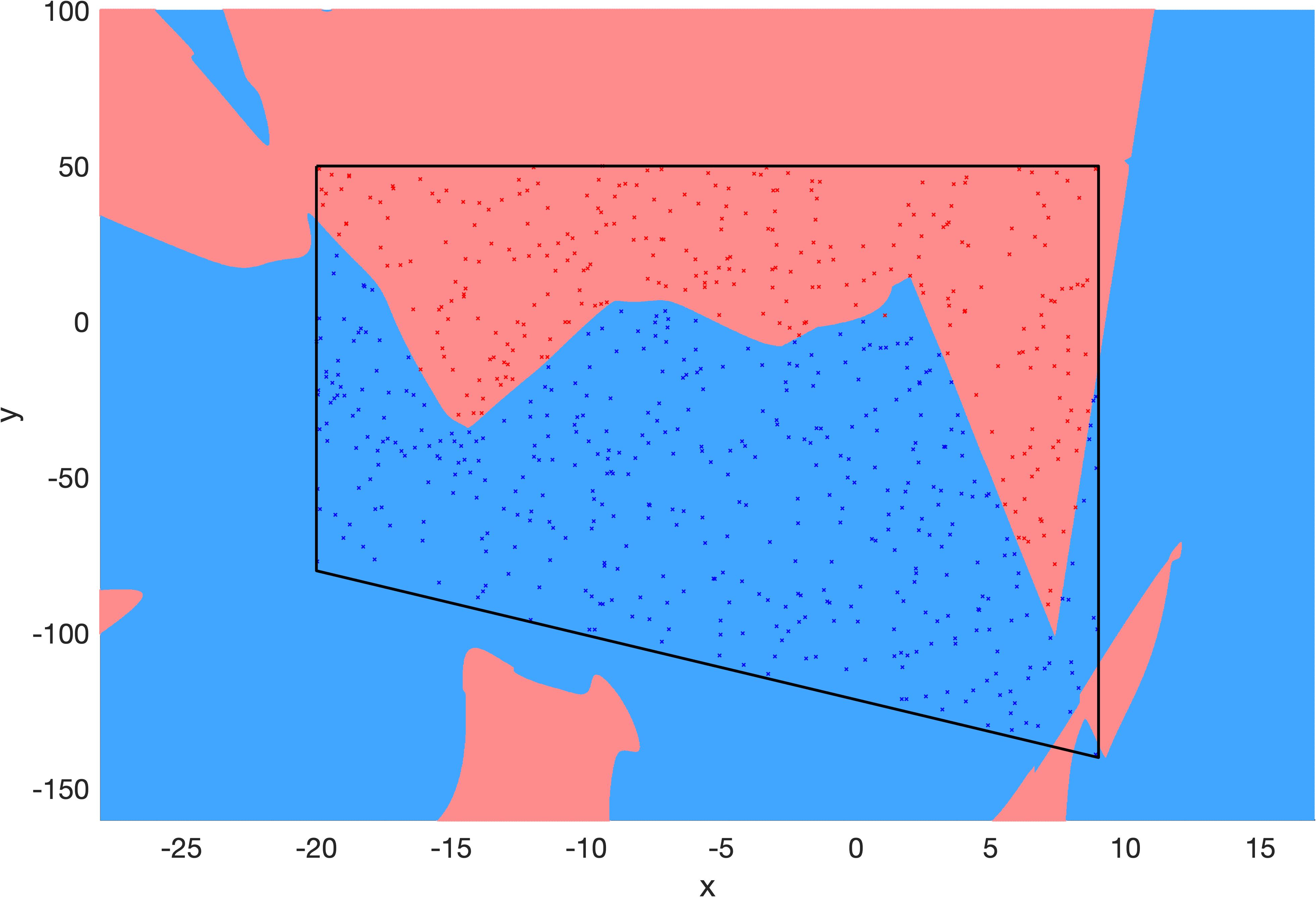}
         \caption{model with 30 neurons}
         \label{fig:nn2}
     \end{subfigure}
    %  \quad
     \begin{subfigure}[b]{0.3\textwidth}
         \centering
         \includegraphics[width=1\linewidth]{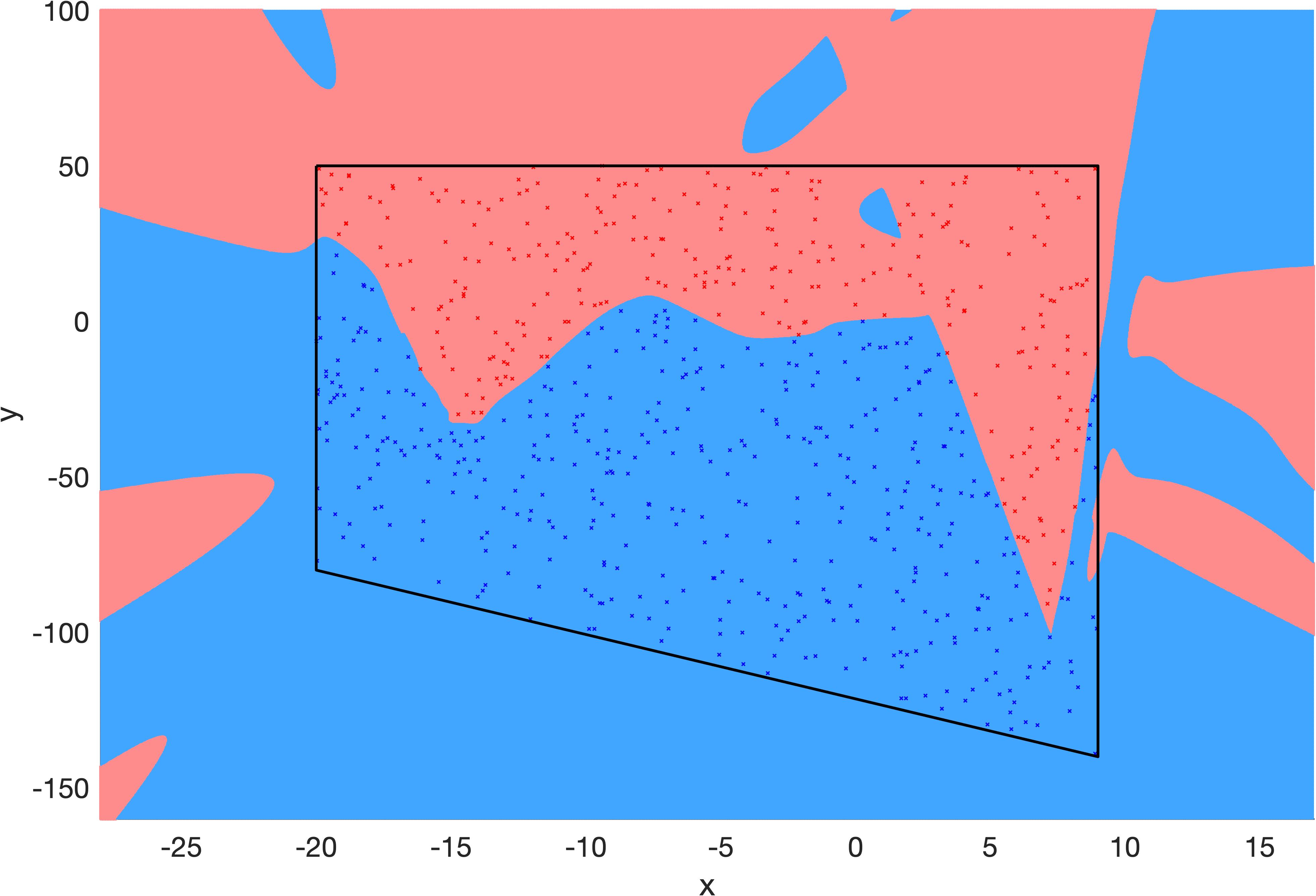}
         \caption{model with 30 neurons}
         \label{fig:nn5}
     \end{subfigure}
    %  \quad
     \begin{subfigure}[b]{0.3\textwidth}
         \centering
         \includegraphics[width=1\linewidth]{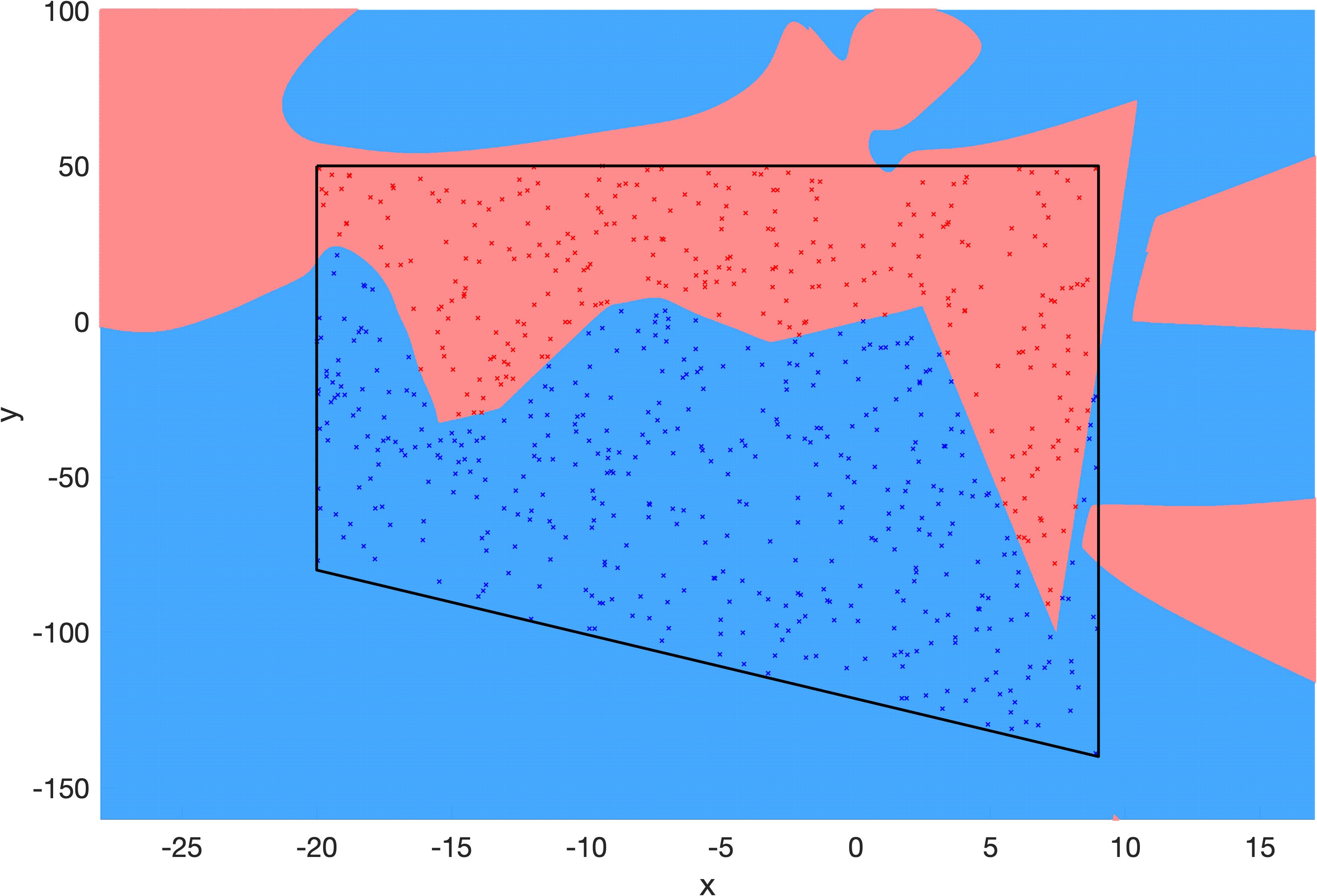}
         \caption{model with 30 neurons}
         \label{fig:nn10}
     \end{subfigure}
     \caption{The case where the model has just about enough capacity to fit the training set. We see that based on the training regime, we can get slightly different decision boundaries inside the hull, all of which perfectly separate the 2 classes. The shape of decision boundaries outside the hull can also vary based on the training regime.}
  \label{fig:model_output_fit}
\end{figure}

\begin{figure}[h]
  \centering
     \begin{subfigure}[b]{0.3\textwidth}
         \centering
         \includegraphics[width=1\linewidth]{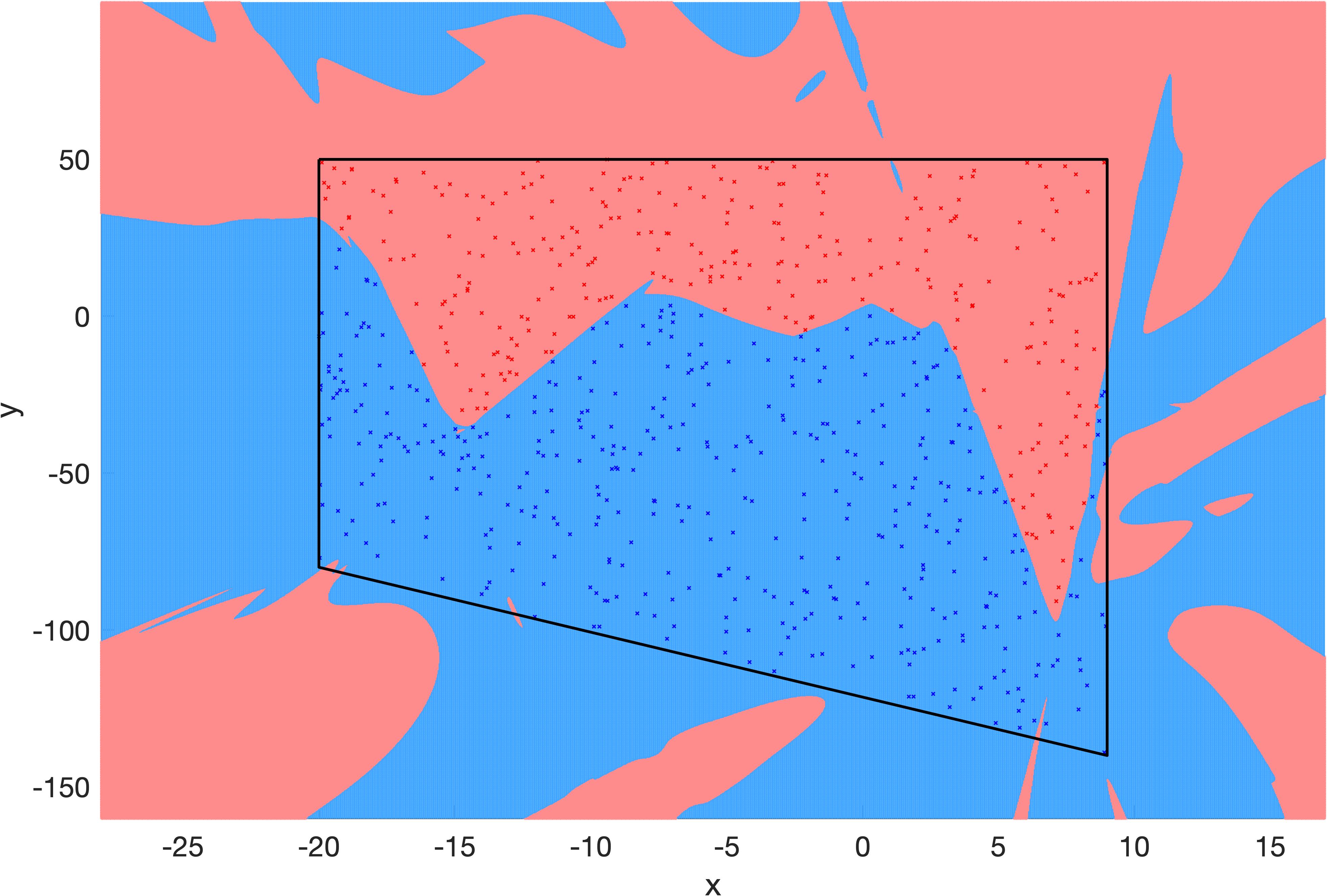}
         \caption{model with 50 neurons}
         \label{fig:nn2}
     \end{subfigure}
    %  \quad
     \begin{subfigure}[b]{0.3\textwidth}
         \centering
         \includegraphics[width=1\linewidth]{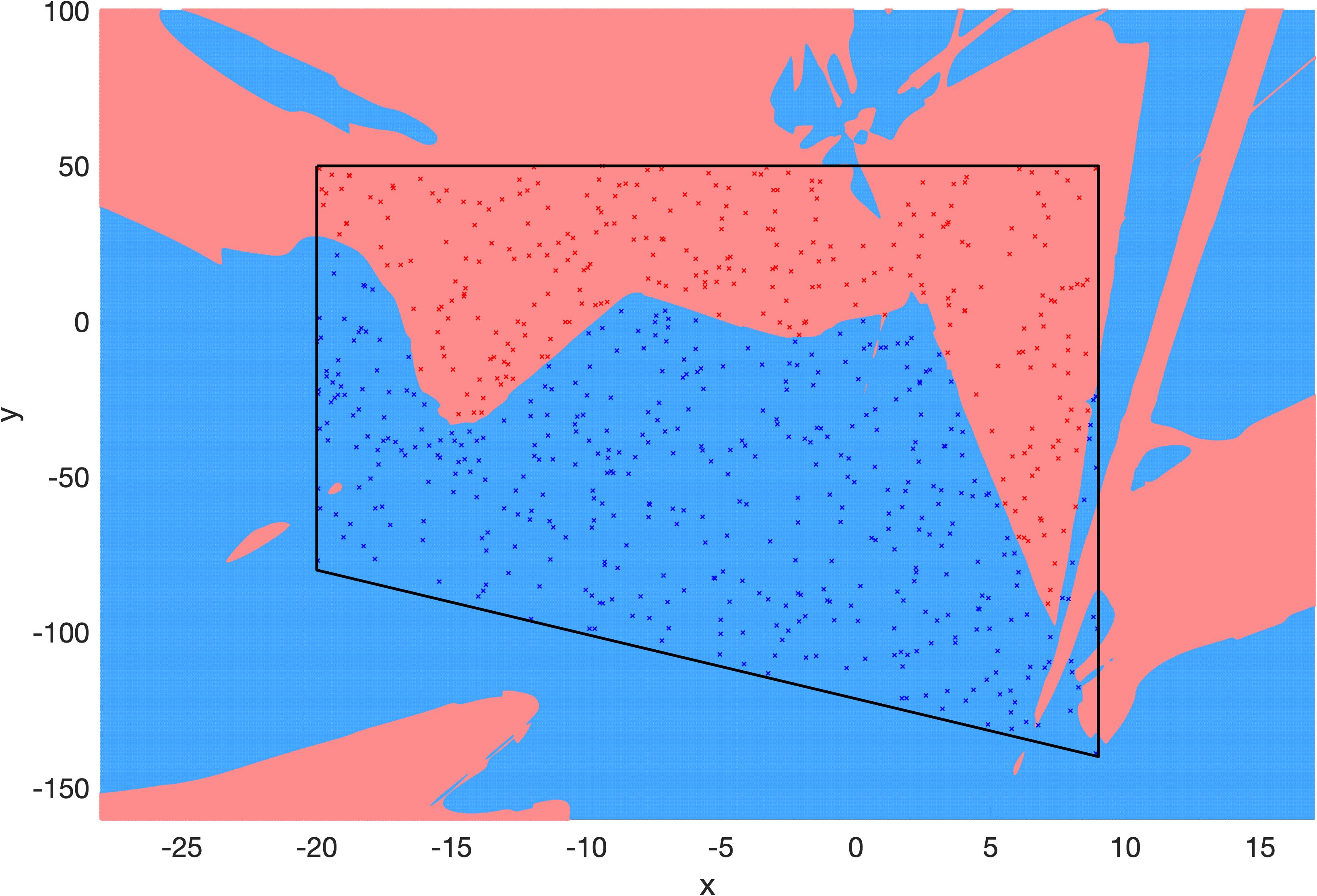}
         \caption{model with 100 neurons}
         \label{fig:nn5}
     \end{subfigure}
    %  \quad
     \begin{subfigure}[b]{0.3\textwidth}
         \centering
         \includegraphics[width=1\linewidth]{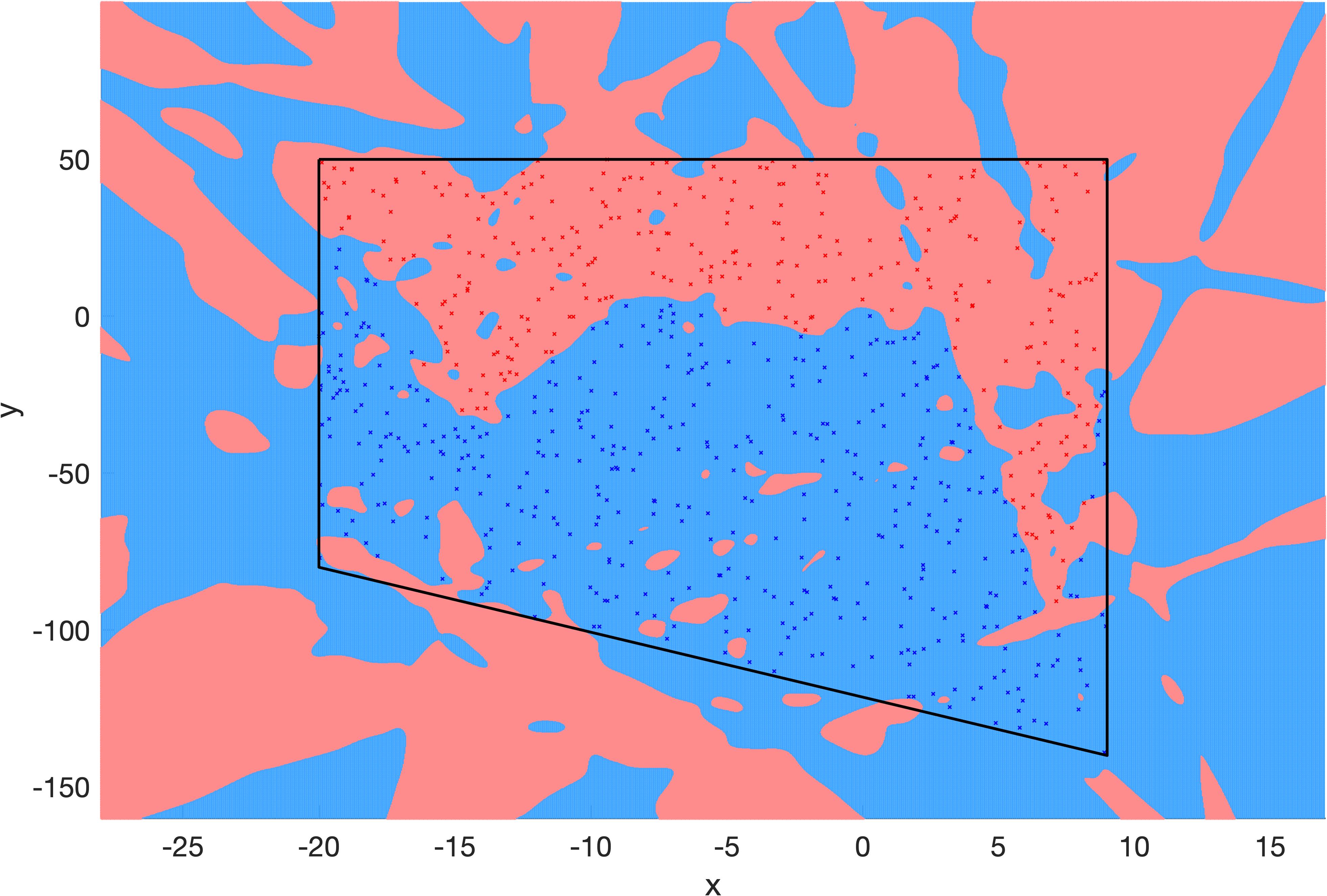}
         \caption{model with 200 neurons}
         \label{fig:nn10}
     \end{subfigure}
    %  \begin{subfigure}[b]{0.3\textwidth}
    %      \centering
    %      \includegraphics[width=1\linewidth]{figures/net_out_1000_1.jpg}
    %      \caption{model with 2 neurons}
    %      \label{fig:nn2}
    %  \end{subfigure}
    % %  \quad
    %  \begin{subfigure}[b]{0.3\textwidth}
    %      \centering
    %      \includegraphics[width=1\linewidth]{figures/net_out_1000_2.jpg}
    %      \caption{model with 5 neurons}
    %      \label{fig:nn5}
    %  \end{subfigure}
    % %  \quad
    %  \begin{subfigure}[b]{0.3\textwidth}
    %      \centering
    %      \includegraphics[width=1\linewidth]{figures/net_out_1000_3.jpg}
    %      \caption{model with 10 neurons}
    %      \label{fig:nn10}
    %  \end{subfigure}
     \caption{As we increase the degree of over-parameterization, from 50 neurons to 200, the number of disjoint decision boundaries increase, inside and outside the hull.}
  \label{fig:model_output_over}
\end{figure}

In this section, we investigate a 2-layer neural network with ReLU activation functions. We train the model with various number of neurons on the data from previous section as depicted in Figure~\ref{fig:polynomial}. We then investigate the output of trained models inside and outside of the $\mathcal{H}^{tr}$, as shown in Figures~\ref{fig:model_output_under}-\ref{fig:model_output_over}. In these figures, the black trapezoid depicts the $\mathcal{H}^{tr}$. The colors red and blue correspond to our 2 classes.

Because of the non-convexity of the loss function, the loss may have numerous minimizers, however, when the model is under-parameterized, as in Figure~\ref{fig:model_output_under}, none of those minimizers would make the loss zero. Finding the same minimizer of training loss is equivalent to obtaining the same trained model, hence unlike the over-parameterized setting, the training regime is focused on finding the global minimizer of the training loss, i.e., finding the best shape for the decision boundary inside the convex hull.

As we increase the number of neurons with increments of 2, we see the model with 30 neurons can perfectly separate the 2 classes in our training set. Figure~\ref{fig:model_output_fit} shows the output of 3 different 30-neuron models that are trained with different training regimes.

Finally, we consider models that are highly over-parameterized. In this regime, there are infinite number of parameter configurations that minimize the training loss to zero, which is equivalent to developing the decision boundaries that perfectly separate our two classes.

The above observations explain why we need over-parameterized models for deep learning and why the generalization of deep learning models are so susceptible to different training regimes. Appendix~\ref{appx:tandem} provides further discussion on this topic and a broader perspective on how interpolation and extrapolation complement each other.
% \begin{enumerate}
%     \item Generalization of deep learning models depends on how they extrapolate;
%     \item In order to desirably shape the extension of decision boundaries, we over-parameterize the models and then minimize their loss using specific training regimes.\footnote{Specific ways of minimizing the training loss imply the massive literature that aim to find the best training regime for achieving the best testing accuracy. Much of that literature has relied on the knowledge of testing/validation sets in order to develop those training methods, i.e., to desirably shape the extension of decision boundaries.}
% \end{enumerate}

\section{Conclusion and future work} \label{sec:conclusion}

We showed that all testing data for standard image classification models considerably lie outside the convex hull of their training sets, in pixel space, in wavelet space, and in the internal representations learned by deep networks. Therefore, the generalization of a deep network relies on its capability to extrapolate outside the boundaries of the data it has seen during training. Based on this observation, the significant number of studies that focus on interpolation regimes seem to be inadequate in explaining the generalization of deep networks.

From this perspective, over-parameterization of models may be considered a necessity to desirably form the extension of decision boundaries outside the convex hull of data. This can be proven for polynomial regression models using the orthogonal system of Legendre polynomials. Moreover, we showed that the training regime can significantly affect the shape of decision boundaries outside the convex hulls, affecting the accuracy of a model in its extrapolation. We investigated a 2-layer ReLU network and a polynomial decision boundary to demonstrate these ideas.

This work opens an avenue to study deep learning models from a new geometric perspective. As we explained earlier, most of the literature on deep learning generalization and their functional behavior is focused on interpolation models and interpolation assumptions. Many studies in the past few years have explicitly used mere interpolation to explain how deep networks learn to classify images and why they need so many parameters. Most notable examples of such studies are \citep{belkin2019reconciling,ma2018power,belkin2018understand}. Our findings would encourage those studies to broaden their geometric perspective and consider extrapolation in tandem with interpolation in order to explain the generalization in deep learning.

------------

\clearpage

\acks{R.Y. thanks Tammy Kolda for helpful discussion and Shai Shalev-Shwartz for a helpful comment. %Analyzing the directions to convex hulls was inspired by a conversation with Misha Belkin.
R.Y. also thanks Yaim Cooper for the discussion that led to Appendix A, and thanks Benjamin Raichel for helpful pointer to approximation algorithms for convex hulls. R.Y. was supported by a fellowship from the Department of Veteran Affairs. The views expressed in this manuscript are those of the author and do not necessarily reflect the position or policy of the Department of Veterans Affairs or the United States government.}

% \clearpage

%\bibliographystyle{abbrvnat}
\bibliography{refs}

\begin{thebibliography}{45}
\providecommand{\natexlab}[1]{#1}
\providecommand{\url}[1]{\texttt{#1}}
\expandafter\ifx\csname urlstyle\endcsname\relax
  \providecommand{\doi}[1]{doi: #1}\else
  \providecommand{\doi}{doi: \begingroup \urlstyle{rm}\Url}\fi

\bibitem[Arora et~al.(2019)Arora, Cohen, Hu, and Luo]{arora2019implicit}
Sanjeev Arora, Nadav Cohen, Wei Hu, and Yuping Luo.
\newblock Implicit regularization in deep matrix factorization.
\newblock In \emph{Advances in Neural Information Processing Systems}, pages
  7413--7424, 2019.

\bibitem[Ascher and Greif(2011)]{ascher2011first}
Uri~M Ascher and Chen Greif.
\newblock \emph{A first course on numerical methods}.
\newblock SIAM, 2011.

\bibitem[Barnard and Wessels(1992)]{barnard1992extrapolation}
Etienne Barnard and LFA Wessels.
\newblock Extrapolation and interpolation in neural network classifiers.
\newblock \emph{IEEE Control Systems Magazine}, 12\penalty0 (5):\penalty0
  50--53, 1992.

\bibitem[Belkin et~al.(2018{\natexlab{a}})Belkin, Hsu, and
  Mitra]{belkin2018overfitting}
Mikhail Belkin, Daniel~J Hsu, and Partha Mitra.
\newblock Overfitting or perfect fitting? {Risk} bounds for classification and
  regression rules that interpolate.
\newblock In \emph{Advances in neural information processing systems}, pages
  2300--2311, 2018{\natexlab{a}}.

\bibitem[Belkin et~al.(2018{\natexlab{b}})Belkin, Ma, and
  Mandal]{belkin2018understand}
Mikhail Belkin, Siyuan Ma, and Soumik Mandal.
\newblock To understand deep learning we need to understand kernel learning.
\newblock In \emph{International Conference on Machine Learning}, pages
  541--549, 2018{\natexlab{b}}.

\bibitem[Belkin et~al.(2019)Belkin, Hsu, Ma, and Mandal]{belkin2019reconciling}
Mikhail Belkin, Daniel Hsu, Siyuan Ma, and Soumik Mandal.
\newblock Reconciling modern machine-learning practice and the classical
  bias--variance trade-off.
\newblock \emph{Proceedings of the National Academy of Sciences}, 116\penalty0
  (32):\penalty0 15849--15854, 2019.

\bibitem[Blum et~al.(2019)Blum, Har-Peled, and Raichel]{blum2019sparse}
Avrim Blum, Sariel Har-Peled, and Benjamin Raichel.
\newblock Sparse approximation via generating point sets.
\newblock \emph{ACM Transactions on Algorithms (TALG)}, 15\penalty0
  (3):\penalty0 1--16, 2019.

\bibitem[Cohen et~al.(2019)Cohen, Rosenfeld, and Kolter]{cohen2019certified}
Jeremy Cohen, Elan Rosenfeld, and Zico Kolter.
\newblock Certified adversarial robustness via randomized smoothing.
\newblock In \emph{International Conference on Machine Learning}, pages
  1310--1320, 2019.

\bibitem[Cohen et~al.(2020)Cohen, Chung, Lee, and
  Sompolinsky]{cohen2020separability}
Uri Cohen, SueYeon Chung, Daniel~D Lee, and Haim Sompolinsky.
\newblock Separability and geometry of object manifolds in deep neural
  networks.
\newblock \emph{Nature communications}, 11\penalty0 (1):\penalty0 1--13, 2020.

\bibitem[Cooper(2018)]{cooper2018loss}
Yaim Cooper.
\newblock The loss landscape of overparameterized neural networks.
\newblock \emph{arXiv preprint arXiv:1804.10200}, 2018.

\bibitem[Daubechies(1992)]{daubechies1992ten}
Ingrid Daubechies.
\newblock \emph{Ten Lectures on Wavelets}.
\newblock Society for Industrial and Applied Mathematics, Philadelphia, 1992.

\bibitem[Ellis et~al.(2018)Ellis, Ritchie, Solar-Lezama, and
  Tenenbaum]{ellis2018learning}
Kevin Ellis, Daniel Ritchie, Armando Solar-Lezama, and Josh Tenenbaum.
\newblock Learning to infer graphics programs from hand-drawn images.
\newblock In \emph{Advances in neural information processing systems}, pages
  6059--6068, 2018.

\bibitem[Elsayed et~al.(2018)Elsayed, Krishnan, Mobahi, Regan, and
  Bengio]{elsayed2018large}
Gamaleldin Elsayed, Dilip Krishnan, Hossein Mobahi, Kevin Regan, and Samy
  Bengio.
\newblock Large margin deep networks for classification.
\newblock In \emph{Advances in Neural Information Processing Systems}, pages
  842--852, 2018.

\bibitem[Fawzi et~al.(2018)Fawzi, Moosavi-Dezfooli, Frossard, and
  Soatto]{fawzi2018empirical}
Alhussein Fawzi, Seyed-Mohsen Moosavi-Dezfooli, Pascal Frossard, and Stefano
  Soatto.
\newblock Empirical study of the topology and geometry of deep networks.
\newblock In \emph{Proceedings of the IEEE Conference on Computer Vision and
  Pattern Recognition}, pages 3762--3770, 2018.

\bibitem[Fink et~al.(2016)Fink, Hershberger, Kumar, and
  Suri]{fink2016hyperplane}
Martin Fink, John Hershberger, Nirman Kumar, and Subhash Suri.
\newblock Hyperplane separability and convexity of probabilistic point sets.
\newblock In \emph{32nd International Symposium on Computational Geometry (SoCG
  2016)}. Schloss Dagstuhl-Leibniz-Zentrum fuer Informatik, 2016.

\bibitem[Goldstein et~al.(2015)Goldstein, Li, and Yuan]{goldstein2015adaptive}
Tom Goldstein, Min Li, and Xiaoming Yuan.
\newblock Adaptive primal-dual splitting methods for statistical learning and
  image processing.
\newblock In \emph{Advances in Neural Information Processing Systems}, pages
  2089--2097, 2015.

\bibitem[Haffner(2002)]{haffner2002escaping}
Patrick Haffner.
\newblock Escaping the convex hull with extrapolated vector machines.
\newblock In \emph{Advances in Neural Information Processing Systems}, pages
  753--760, 2002.

\bibitem[He et~al.(2016)He, Zhang, Ren, and Sun]{he2016deep}
Kaiming He, Xiangyu Zhang, Shaoqing Ren, and Jian Sun.
\newblock Deep residual learning for image recognition.
\newblock In \emph{Proceedings of the IEEE conference on computer vision and
  pattern recognition}, pages 770--778, 2016.

\bibitem[Hueter(1999)]{hueter1999limit}
Irene Hueter.
\newblock Limit theorems for the convex hull of random points in higher
  dimensions.
\newblock \emph{Transactions of the American Mathematical Society},
  351\penalty0 (11):\penalty0 4337--4363, 1999.

\bibitem[Jiang et~al.(2019)Jiang, Krishnan, Mobahi, and
  Bengio]{marginbased2019}
Yiding Jiang, Dilip Krishnan, Hossein Mobahi, and Samy Bengio.
\newblock Predicting the generalization gap in deep networks with margin
  distributions.
\newblock In \emph{International Conference on Learning Representations}, 2019.

\bibitem[Kanbak et~al.(2018)Kanbak, Moosavi-Dezfooli, and
  Frossard]{kanbak2018geometric}
Can Kanbak, Seyed-Mohsen Moosavi-Dezfooli, and Pascal Frossard.
\newblock Geometric robustness of deep networks: Analysis and improvement.
\newblock In \emph{Proceedings of the IEEE Conference on Computer Vision and
  Pattern Recognition}, pages 4441--4449, 2018.

\bibitem[Kileel et~al.(2019)Kileel, Trager, and Bruna]{kileel2019expressive}
Joe Kileel, Matthew Trager, and Joan Bruna.
\newblock On the expressive power of deep polynomial neural networks.
\newblock In \emph{Advances in Neural Information Processing Systems}, pages
  10310--10319, 2019.

\bibitem[Kosanovich et~al.(1996)Kosanovich, Gurumoorthy, Sinzinger, and
  Piovoso]{kosanovich1996improving}
K~Kosanovich, A~Gurumoorthy, E~Sinzinger, and M~Piovoso.
\newblock Improving the extrapolation capability of neural networks.
\newblock In \emph{Proceedings of the 1996 IEEE International Symposium on
  Intelligent Control}, pages 390--395. IEEE, 1996.

\bibitem[Krizhevsky(2009)]{krizhevsky2009learning}
Alex Krizhevsky.
\newblock Learning multiple layers of features from tiny images.
\newblock 2009.

\bibitem[LeCun et~al.(1998)LeCun, Bottou, Bengio, and
  Haffner]{lecun1998gradient}
Yann LeCun, L{\'e}on Bottou, Yoshua Bengio, and Patrick Haffner.
\newblock Gradient-based learning applied to document recognition.
\newblock \emph{Proceedings of the IEEE}, 86\penalty0 (11):\penalty0
  2278--2324, 1998.

\bibitem[Liang et~al.(2020)Liang, Rakhlin, et~al.]{liang2020just}
Tengyuan Liang, Alexander Rakhlin, et~al.
\newblock Just interpolate: Kernel “ridgeless” regression can generalize.
\newblock \emph{Annals of Statistics}, 48\penalty0 (3):\penalty0 1329--1347,
  2020.

\bibitem[Ma et~al.(2018)Ma, Bassily, and Belkin]{ma2018power}
Siyuan Ma, Raef Bassily, and Mikhail Belkin.
\newblock The power of interpolation: Understanding the effectiveness of sgd in
  modern over-parametrized learning.
\newblock In \emph{International Conference on Machine Learning}, pages
  3325--3334. PMLR, 2018.

\bibitem[Neyshabur et~al.(2017{\natexlab{a}})Neyshabur, Bhojanapalli,
  McAllester, and Srebro]{neyshabur2017exploring}
Behnam Neyshabur, Srinadh Bhojanapalli, David McAllester, and Nati Srebro.
\newblock Exploring generalization in deep learning.
\newblock In \emph{Advances in Neural Information Processing Systems}, pages
  5947--5956, 2017{\natexlab{a}}.

\bibitem[Neyshabur et~al.(2017{\natexlab{b}})Neyshabur, Tomioka, Salakhutdinov,
  and Srebro]{neyshabur2017geometry}
Behnam Neyshabur, Ryota Tomioka, Ruslan Salakhutdinov, and Nathan Srebro.
\newblock Geometry of optimization and implicit regularization in deep
  learning.
\newblock \emph{arXiv preprint arXiv:1705.03071}, 2017{\natexlab{b}}.

\bibitem[Neyshabur et~al.(2019)Neyshabur, Li, Bhojanapalli, LeCun, and
  Srebro]{neyshabur2019towards}
Behnam Neyshabur, Zhiyuan Li, Srinadh Bhojanapalli, Yann LeCun, and Nathan
  Srebro.
\newblock Towards understanding the role of over-parametrization in
  generalization of neural networks.
\newblock In \emph{International Conference on Learning Representations}, 2019.

\bibitem[{Nocedal} and {Wright}(2006)]{nocedal2006numerical}
Jorge {Nocedal} and Stephen {Wright}.
\newblock \emph{Numerical Optimization}.
\newblock Springer, New York, 2nd edition, 2006.
\newblock ISBN 9780387400655.

\bibitem[{O'Leary} and Rust(2013)]{oleary2013variable}
Dianne~P {O'Leary} and Bert~W Rust.
\newblock Variable projection for nonlinear least squares problems.
\newblock \emph{Computational Optimization and Applications}, 54\penalty0
  (3):\penalty0 579--593, 2013.

\bibitem[Papernot et~al.(2016)Papernot, McDaniel, Jha, Fredrikson, Celik, and
  Swami]{papernot2016limitations}
Nicolas Papernot, Patrick McDaniel, Somesh Jha, Matt Fredrikson, Z~Berkay
  Celik, and Ananthram Swami.
\newblock The limitations of deep learning in adversarial settings.
\newblock In \emph{IEEE European symposium on security and privacy (EuroS\&P)},
  pages 372--387. IEEE, 2016.

\bibitem[Psichogios and Ungar(1992)]{psichogios1992hybrid}
Dimitris~C Psichogios and Lyle~H Ungar.
\newblock A hybrid neural network-first principles approach to process
  modeling.
\newblock \emph{AIChE Journal}, 38\penalty0 (10):\penalty0 1499--1511, 1992.

\bibitem[Savarese et~al.(2019)Savarese, Evron, Soudry, and
  Srebro]{savarese2019infinite}
Pedro Savarese, Itay Evron, Daniel Soudry, and Nathan Srebro.
\newblock How do infinite width bounded norm networks look in function space?
\newblock In \emph{Conference on Learning Theory}, pages 2667--2690, 2019.

\bibitem[Shafahi et~al.(2019)Shafahi, Huang, Studer, Feizi, and
  Goldstein]{shafahi2018adversarial}
Ali Shafahi, W~Ronny Huang, Christoph Studer, Soheil Feizi, and Tom Goldstein.
\newblock Are adversarial examples inevitable?
\newblock In \emph{International Conference on Learning Representations}, 2019.

\bibitem[Snell et~al.(2017)Snell, Swersky, and Zemel]{snell2017prototypical}
Jake Snell, Kevin Swersky, and Richard Zemel.
\newblock Prototypical networks for few-shot learning.
\newblock In \emph{{NeurIPS}}, pages 4077--4087, 2017.

\bibitem[Strang(2019)]{strang2019linear}
Gilbert Strang.
\newblock \emph{Linear Algebra and Learning from Data}.
\newblock Wellesley-Cambridge Press, 2019.

\bibitem[Tsipras et~al.(2019)Tsipras, Santurkar, Engstrom, Turner, and
  Madry]{tsipras2018robustness}
Dimitris Tsipras, Shibani Santurkar, Logan Engstrom, Alexander Turner, and
  Aleksander Madry.
\newblock Robustness may be at odds with accuracy.
\newblock In \emph{International Conference on Learning Representations}, 2019.

\bibitem[Verma et~al.(2019)Verma, Lamb, Beckham, Najafi, Mitliagkas, Lopez-Paz,
  and Bengio]{verma2019manifold}
Vikas Verma, Alex Lamb, Christopher Beckham, Amir Najafi, Ioannis Mitliagkas,
  David Lopez-Paz, and Yoshua Bengio.
\newblock Manifold mixup: Better representations by interpolating hidden
  states.
\newblock In \emph{International Conference on Machine Learning}, pages
  6438--6447. PMLR, 2019.

\bibitem[Vincent and Bengio(2002)]{vincent2002k}
Pascal Vincent and Yoshua Bengio.
\newblock K-local hyperplane and convex distance nearest neighbor algorithms.
\newblock In \emph{Advances in neural information processing systems}, pages
  985--992, 2002.

\bibitem[Xiao et~al.(2020)Xiao, Engstrom, Ilyas, and Madry]{xiao2020noise}
Kai Xiao, Logan Engstrom, Andrew Ilyas, and Aleksander Madry.
\newblock Noise or signal: The role of image backgrounds in object recognition.
\newblock \emph{arXiv preprint arXiv:2006.09994}, 2020.

\bibitem[Xu et~al.(2020)Xu, Li, Zhang, Du, Kawarabayashi, and
  Jegelka]{xu2020neural}
Keyulu Xu, Jingling Li, Mozhi Zhang, Simon~S Du, Ken-ichi Kawarabayashi, and
  Stefanie Jegelka.
\newblock How neural networks extrapolate: From feedforward to graph neural
  networks.
\newblock \emph{arXiv preprint arXiv:2009.11848}, 2020.

\bibitem[Yousefzadeh and Huang(2020)]{yousefzadeh2020wspectral}
Roozbeh Yousefzadeh and Furong Huang.
\newblock Using wavelets and spectral methods to study patterns in
  image-classification datasets.
\newblock \emph{arXiv preprint arXiv:2006.09879}, 2020.

\bibitem[Zhang et~al.(2017)Zhang, Bengio, Hardt, Recht, and
  Vinyals]{zhang2016understanding}
Chiyuan Zhang, Samy Bengio, Moritz Hardt, Benjamin Recht, and Oriol Vinyals.
\newblock Understanding deep learning requires rethinking generalization.
\newblock In \emph{International Conference on Learning Representations}, 2017.

\end{thebibliography}

\clearpage
\appendix

\section{The case of random data in high-dimensional space} \label{appx:random_points}

The question may arise that what happens if we have random data points in the same high-dimensional space. Would the data points still fall outside the \cvx? Would the distance to \cvx be the same? The answers are yes and no, respectively. 

To gain some insights, let's consider the CIFAR-10 dataset. We can randomly shuffle all the pixel values in all the images of training and testing sets. In such case, the shuffled testing data would still fall outside the \cvx of shuffled data. But, their distance to \cvx would be orders of magnitude larger. 

Alternatively, when we generate random points in the same domain as the pixel space of CIFAR-10, (i.e., domain with 3,072 dimensions bounded between 0 and 255), again, the testing data are outside the \cvx.\footnote{This relates to the limit theorems for the convex hull of random points in higher dimensions \citep{hueter1999limit} and also to studies on separability and distribution of random points \citep{fink2016hyperplane}.} This time, the distances to \cvx are much larger even compared to the case of shuffling the pixel values. 

We can conclude that with such number of training samples in such high-dimensional domains, one can expect the testing samples to be outside their \cvx. However, the distance to the convex hulls are much closer for our image datasets, compared to random data points, because for each testing image, $x^{te}$, there are a group of training images that form a surface on the \cvx, and that surface is much closer to $x^{te}$, compared to the surface that a set of random data points can create.% In Figures~\ref{fig:conv_perturb_cif10} and~\ref{fig:conv_perturb_mnist}, we showed that in fact, the point on the \cvx, closest to an image, depicts a vague demonstration of that same image.

As we showed earlier, the directions to \cvx contain information about the objects of interest depicted in images. Hence, the extrapolation task of a neural network is meaningful and related to what it needs to classify in images. Recently, there has been further studies that try to learn images from even fewer training samples, i.e., few-shot learning \citep{snell2017prototypical}, which implies a more advanced extrapolation task. This is also pursued by recent studies in cognitive science \citep{ellis2018learning}.

\setcounter{figure}{0}
\setcounter{equation}{0}
\renewcommand{\thefigure}{B\arabic{figure}}

\section{Extrapolation and interpolation work in tandem} \label{appx:tandem}

Earlier, we reported that the extent of extrapolation, for the CIFAR-10 dataset, is about 27\% of the diameter of its \cvx. We also reported that all testing samples are outside the \cvx. Based on these observations, we can say that the task of classifying the testing set of CIFAR-10 is extrapolation. But, how does the \textbf{interpolation} affect our ability to \textbf{extrapolate}? We can explain this using the decision boundaries of the models.

For image classification, our domain is a $d$-dimensional hyper-cube, because pixel values are bounded. $d$ is the number of pixels. The convex hull of training set, \cvx, is a shape with $\leq d$ dimension, and sits somewhere in that hyper-cube. For CIFAR-10 dataset, \cvx has exactly $d$ dimensions. The testing samples sit around the \cvx, like a mist, not too far, and not too close to \cvx. The distance of testing samples from \cvx ranges between 1\%-27\% of its diameter, with average value of 10\%. 

As we mentioned earlier, a classification function/model partitions its domain and assigns a class to each of the partitions. The partitions are defined by decision boundaries, and so is the model. Basically, the training process partitions the shape \cvx by defining a finite set of decision boundaries inside it. We can define this process as interpolation, especially if we are only concerned about the location of decision boundaries in \cvx. Some of the decision boundaries defined in this process will reach the surface of \cvx and extend outside it. Theses decision boundaries and their extensions outside the hull are the ones that a model relies upon in order to classify the images outside the hull.

Because the testing images are not too far outside the hull, the space to shape the extensions of decision boundaries is limited. Therefore, the locations where the decision boundaries reach the surface of \cvx is of great importance. Overall, interpolation and extrapolation work in tandem to shape the decision boundaries and the functional performance of image classification models.

Going back to the hyper-cube, during the training, the shape of \cvx is partitioned with some nonlinear surfaces (decision boundaries), and some of those surfaces extend outside the \cvx. The locations where the decision boundaries reach the surface of \cvx and their extension outside the \cvx is critical in how the model classifies the testing images that are sitting around the \cvx.

% Figure~\ref{fig:2d_domain_cvx} tries to depict this view in a simple imaginary example in a 2D domain.

% \begin{figure}[H]
%     \centering
%      \begin{subfigure}[b]{0.35\textwidth}
%          \centering
%          \includegraphics[width=1\linewidth]{figures/2d_domain_cvx.jpg}
%         %  \caption{MNIST (pixel space)}
%         %  \label{fig:data}
%      \end{subfigure}
%     \caption{A simple imaginary example of a bounded domain in 2D,  partitioned with decision boundaries. The \cvx is depicted with dark blue, and a set of testing samples surround it.}
%   \label{fig:2d_domain_cvx}
% \end{figure}


\newpage

% \vskip 0.2in
% \bibliography{refs}

\end{document}